\documentclass[runningheads]{llncs}

\usepackage{eccv}

\usepackage{eccvabbrv}

\usepackage{graphicx}
\usepackage{booktabs}
\usepackage[accsupp]{axessibility}  
\usepackage{subcaption} 
\usepackage{tikz}
\usepackage{comment}
\usepackage{amsmath,amssymb} 
\usepackage{color}
\usepackage{soul}
\usepackage[normalem]{ulem}
\usepackage{multirow}

\usepackage{orcidlink}
\useunder{\uline}{\ul}{}

\begin{document}

\title{Enhanced Online Test-time Adaptation with Feature-Weight Cosine Alignment} 



\author{WeiQin Chuah\orcidlink{0000-0001-6020-4660} \and
Ruwan Tennakoon\orcidlink{0000-0001-8909-5728} \and
Alireza Bab-Hadiashar\orcidlink{0000-0002-6192-2303}}

\authorrunning{W.~Chuah et al.}

\institute{
RMIT University, Melbourne Australia \\ 
\email{\{wei.qin.chuah, ruwan.tennakoon, alireza.bab-hadiashar\}@rmit.edu.au}}

\maketitle

\begin{abstract}
    Online Test-Time Adaptation (OTTA) has emerged as an effective strategy to handle distributional shifts, allowing on-the-fly adaptation of pre-trained models to new target domains during inference, without the need for source data. We uncovered that the widely studied entropy minimization (EM) method for OTTA, suffers from noisy gradients due to ambiguity near decision boundaries and incorrect low-entropy predictions. To overcome these limitations, this paper introduces a novel cosine alignment optimization approach with a dual-objective loss function that refines the precision of class predictions and adaptability to novel domains. Specifically, our method optimizes the cosine similarity between feature vectors and class weight vectors, enhancing the precision of class predictions and the model's adaptability to novel domains. Our method outperforms state-of-the-art techniques and sets a new benchmark in multiple datasets, including CIFAR-10-C, CIFAR-100-C, ImageNet-C, Office-Home, and DomainNet datasets, demonstrating high accuracy and robustness against diverse corruptions and domain shifts. The code and implementation details are available on our official \href{https://github.com/waychin-weiqin/CoMM}{GitHub repository}.
\end{abstract}

\section{Introduction}
\label{sec:intro}
Deep learning advancements have significantly improved visual recognition capabilities, yet these models often underperform when the test data deviates slightly from the training distribution due to factors like changes in lighting conditions or the presence of data corruption.  Conventionally, overcoming these shifts required access to either unlabeled target data during the training phase or source data from diverse domains, as seen in domain adaptation~\cite{baktashmotlagh2013unsupervised,ganin2015unsupervised,saito2018maximum} and generalization~\cite{li2018learning,li2018domain,zhou2022domain} strategies. 

\begin{figure}[t]
\centering
\begin{subfigure}[t]{0.4\textwidth}
    \includegraphics[width=\textwidth]{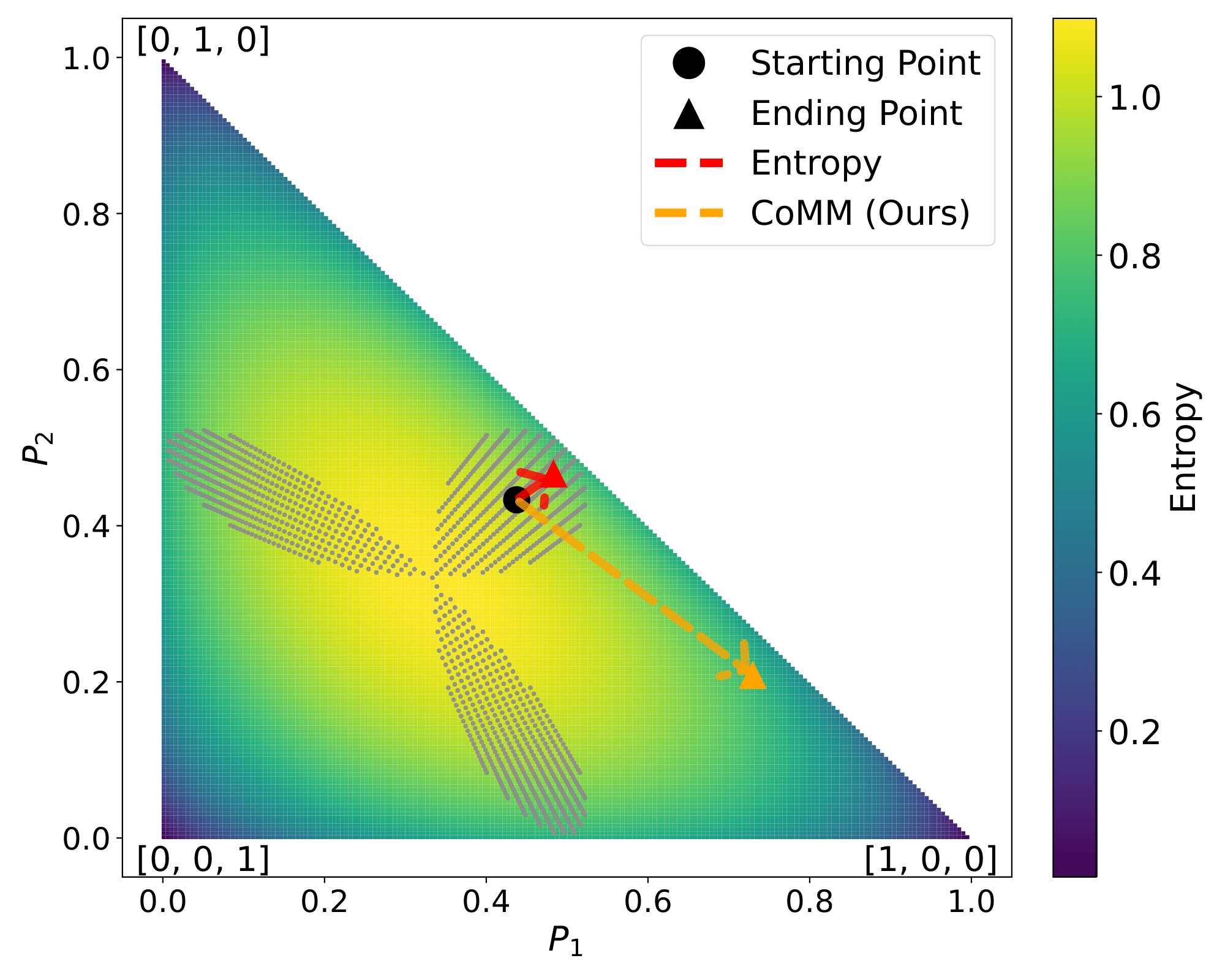}
    \caption{}
    \label{fig:simplex}
\end{subfigure}
\hfill
\begin{subfigure}[t]{0.57\textwidth}
    \includegraphics[width=\textwidth]{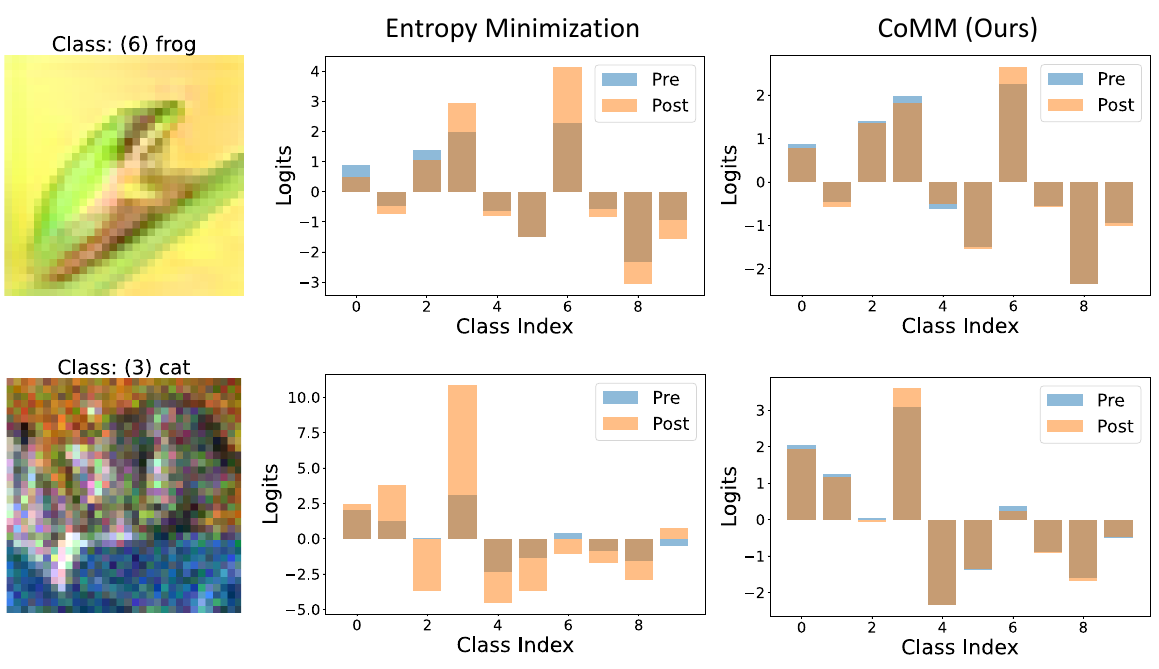}
    \caption{}
    \label{fig:frog}
\end{subfigure}
\caption{Analytical comparison between entropy minimization (EM) and our proposed method (CoMM) in terms of optimization. Figure (a) illustrates the EM loss function within the probability simplex for a three-class toy problem, including optimization trajectories. For initialization within specific regions (highlighted in gray), EM optimization steers predictions towards a lower-entropy area. However, this process results in at least two logits having positive gradients, causing the model to align with multiple classes and leading to confusion. Figure (b) demonstrates EM's limitations using CIFAR-10-C dataset examples: The top row shows EM increasing logits for both the correct class 6 (frog) and an incorrect class 3 (cat), leading to ambiguity. Similarly, the bottom row indicates EM raising logits for the correct class 3 (cat) and incorrectly for classes 0 (airplane) and 1 (automobile). In contrast, our proposed CoMM method shows a more targeted reduction in entropy, indicative of decisive classification.}
\label{fig:motivation_low_margin}
\vspace{-5pt}
\end{figure}

\begin{figure}[t]
\centering
\begin{subfigure}[b]{0.4\textwidth}
    \includegraphics[width=\textwidth]{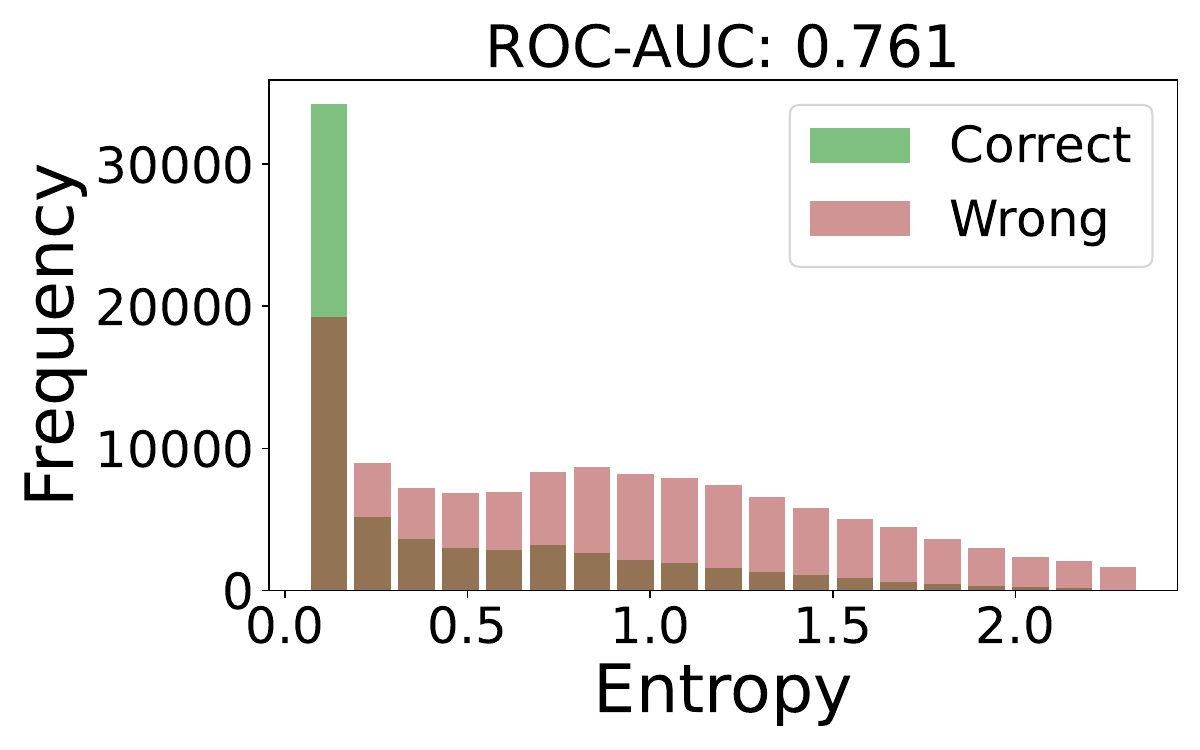}
\end{subfigure}
\begin{subfigure}[b]{0.4\textwidth}
    \includegraphics[width=\textwidth]{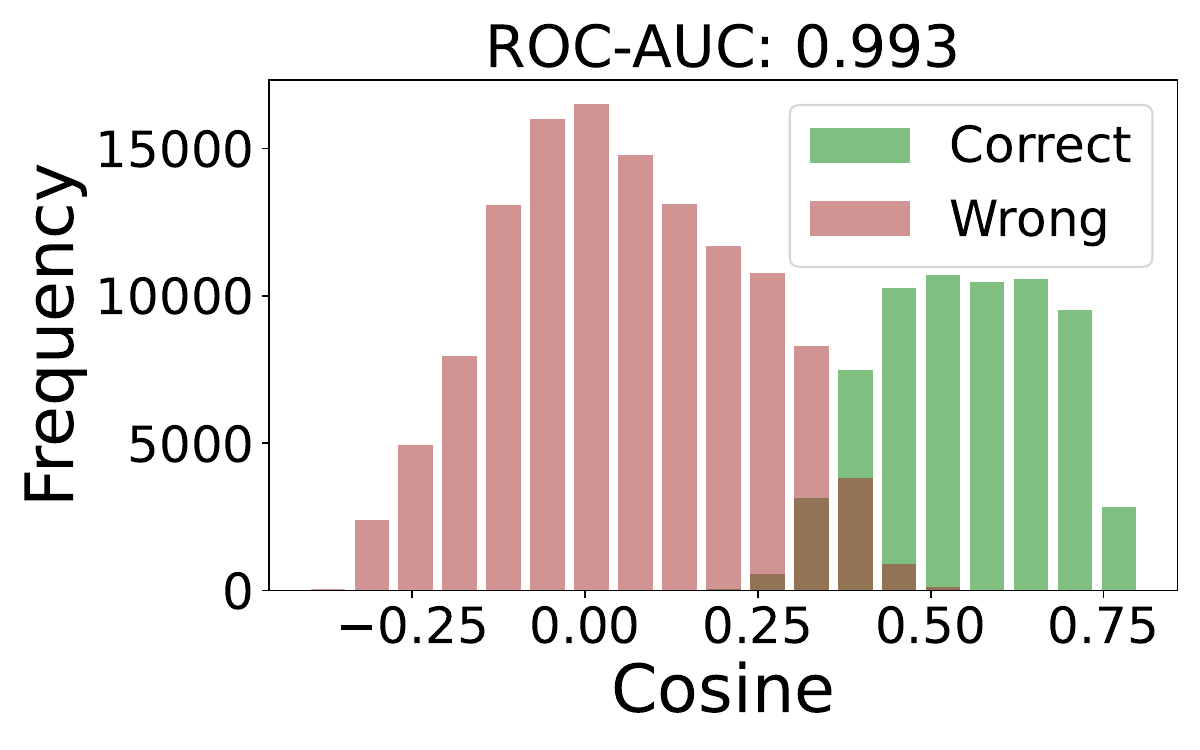}
\end{subfigure}
\caption{Distribution of entropy (left) and cosine similarity (right) for correctly (green) and incorrectly (red) classified out-of-domain samples in CIFAR-10-C across all corruption types at severity level 5. These distributions provide insights into the discriminative power of the model's predictions, reflecting the impact of data corruption on prediction certainty and feature-class alignment.}
\label{fig:motivation}
\end{figure}

Recently, Test-Time Adaptation (TTA) has emerged as a promising approach, enabling pre-trained models to adapt to new, unseen data distributions without requiring labeled data at inference time~\cite{liang2023comprehensive,wang2023search}. A practical paradigm of TTA is the online test-time adaptation (OTTA) problem, where the pre-trained model is adapted to a mini-batch of target data that can only be observed once~\cite{liang2023comprehensive}. Among the methodologies proposed for OTTA~\cite{iwasawa2021test,mirza2022norm,wang2021tent,kojima2022robustifying}, the entropy minimization (EM) method, which was proposed by Wang~\etal~\cite{wang2021tent}, is extensively explored and served as the basis approach for many follow-up works~\cite{yi2023temporal,tang2023neuro,park2023robust,fleuret2021test,zhou2021bayesian}. By utilizing EM as a surrogate objective, these approaches employ predictive certainty derived from entropy as an implicit supervisory signal, enabling models to adjust to new data distribution without ground-truth labels. This approach has been proven to substantially enhance OTTA performance across a variety of benchmarks, marking a significant advancement in the field.
Despite the impressive results, we found that the EM method is susceptible to:
\begin{enumerate}
    \item \textit{Gradient Ambiguity Near Decision Boundaries}: The EM loss function, when applied to samples that are near decision boundaries, exhibits an interesting behavior. Specifically, the gradient with respect to logits for those samples would contain more than one positive element. As a consequence of utilizing the EM loss function for model adaptation during test time, the feature extractor is optimized for multiple classes based on a single sample, thereby introducing confusion by simultaneously aligning the extracted features with several classes. Figure~\ref{fig:simplex}, provides an illustrative example. The figure depicts the entropy loss function on the probability simplex for a toy problem with three classes and the optimization trajectories. For points within certain regions close to the boundary, i.e., the region depicted in gray overlay, optimizing the EM loss function moves the predictions towards a lower-entropy region, but the gradients for logits will have at least two positive elements causing the features to alight more with both classes creating confusion. A real example from CIFAR-10-C dataset is shown in Figure~\ref{fig:frog}, where the EM method increases the logit scores for both the correct class (6: frog) and an incorrect class (3: cat), instead of selectively increasing the logit score of the predicted class only.
    
    \item \textit{Overconfident predictions for Out-of-domain data}: As shown in Figure~\ref{fig:motivation} (left), the predictions with low entropy (those close to 0) can often be incorrect\footnote{This phenomenon aligns with the known issue of overconfidence bias in machine learning~\cite{nguyen2015deep}.}, especially for out-of-distribution test data. This finding challenges the common thinking that low entropy is a reliable measure of prediction correctness~\cite{wang2021tent}. As such, relying exclusively on EM loss function for OTTA can inadvertently increase the model's confidence in incorrect predictions, creating an illusion of successful adaptation. Furthermore, many samples (those close to 0 entropy) would not yield useful learning signals (i.e., gradients), and not participate in the training process. This is particularly undesirable for OTTA where a sample is presented to the learning algorithm only once; If not used, that sample would never have an opportunity to influence the learning process. 
\end{enumerate}

Motivated by the above findings, we aim to identify a more effective surrogate loss than EM loss function for the OTTA. To this end, we leverage the concept of cosine similarity, which measures the cosine of the angle between the extracted features and classifier weight vectors. In contrast to the prediction's entropy, we found that emphasizing the alignment of cosine similarity for OTTA provides a more precise and targeted adaptation of the source model to the target data. Specifically, this method enables direct one-to-one alignment, similar to the specificity offered by cross-entropy, in contrast to the less constrained approach of the EM method that often struggles with \textit{gradient ambiguity}. Moreover, as shown in Figure~\ref{fig:motivation}~(right), higher cosine similarity values are a better indicator of correct vs incorrect predictions in comparison with entropy. 
Our cosine alignment objective is designed to increase the cosine similarity between the feature vector of a target instance and the weight vector of its predicted class, while concurrently decreasing its similarity with the weight vectors of non-predicted classes. This dual objective aims to enhance both the precision and specificity of the adaptation process, significantly improving model performance in the test domain. As we show in Figure~\ref{fig:motivation_low_margin}, our proposed method can effectively mitigate the limitations of the EM method, thus achieving significant performance improvement in OTTA across multiple benchmarks. To summarize, the main contributions of our work are as follows:
\begin{itemize}
    \item The limitations of the commonly used entropy minimization method in handling samples near the boundaries and overconfident predictions during OTTA are identified.
    \item An innovative unsupervised loss function based on feature-weight cosine similarity, enhancing OTTA performance by aligning target data representations with source classifier weights is developed. This is shown to improve model robustness and adaptability to distributional shifts in unseen data
    \item Our method achieves state-of-the-art results on a wide array of benchmarks, including CIFAR-10-C, CIFAR-100-C, ImageNet-C, OfficeHome, and DomainNet, establishing new standards for OTTA.
\end{itemize}

\section{Related  Work}
\textbf{Test-time Adaptation}: Test-Time Adaptation (TTA) aims to enhance model robustness by adapting pre-trained models to new data distributions encountered during inference, without revisiting the original training data. This setup is particularly useful when the test data distribution deviates from the source data distribution.  Another closely related approach, Test-time Training (TTT)~\cite{liu2021ttt++,sun2020test}, employs self-supervised tasks such as rotation prediction or contrastive loss to aid adaptation, often requiring retraining with the auxiliary task added to the original model. However, the relevance of these proxy tasks to the primary classification goal is not guaranteed, which may lead to sub-optimal task-specific performance improvements~\cite{wang2021tent}. 
In contrast to TTT, Online Test-Time Adaptation (OTTA) methods continuously updates the model on-the-fly as it encounters new data during inference, ensuring immediate adaptation to evolving data distributions while bypassing the need to modify the original training regimen~\cite{chen2022contrastive,kojima2022robustifying,liang2020we,mirza2023actmad,nguyen2023tipi,wang2021tent,zhang2022memo}. Notably, TENT~\cite{wang2021tent} specifically addresses distributional shifts by dynamically adjusting batch normalization parameters using entropy loss minimization during inference. Similarly, EATA~\cite{niu2022efficient} introduces a selective approach to optimizing unsupervised surrogate losses like TENT's by focusing only on reliable and informative data points. 

In this work, we show that methods relying on the entropy loss minimization approach (e.g. TENT) are susceptible to noisy gradients from overconfident or low-margin predictions. To overcome this limitation, we propose an unsupervised loss function based on feature-weight cosine similarity. The proposed approach can effectively address the issues of noisy gradients and align the target data representations more accurately with the source classifier weights, leading to more reliable and robust performance. Our technique, therefore, bridges the gap between the need for domain adaptability and the maintenance of high classification accuracy, setting new standards in OTTA across a variety of benchmark datasets. 
\vspace{5pt}

\noindent \textbf{Leveraging Cosine Similarity for Improved Model Performance}: 
The role of cosine similarity in enhancing model performance has been a subject of interest across various domains of deep learning, such as generalization and explainability. 
For instance, {Liu~\etal}~\cite{liu2018decoupled} introduced a decoupled learning strategy within convolutional neural networks (CNNs), distinguishing the learning of feature magnitudes—responsible for intra-class variation—from feature orientations, which discern semantic differences. This decoupling provides deeper insight into the model's decision-making processes. 
Extending this concept, {Chuah~\etal}~\cite{chuah2024single} proposed a centered cosine similarity technique to improve single source domain generalization, reducing the dependence on data augmentation strategies. In a different direction, {Böhle~\etal}~\cite{bohle2022b} highlights the importance of feature-weight alignment in the interpretative process. They introduced the B-cos transformation, a novel modification of the conventional linear transformation in neural networks, which increases weight-input alignment during optimization. Their findings demonstrate that such an alignment-centric approach can substantially elevate a model's interpretability, suggesting that this method could offer a valuable perspective on the internal workings of neural computations.

In contrast, this work focuses on the application of cosine similarity in test-time adaptation, proposing an unsupervised loss function that not only aligns the target and source domains more accurately but also enhances model adaptability in the face of distributional changes, as evidenced by our state-of-the-art results on multiple benchmarks.

\section{Proposed Method}
\subsection{Problem Setup}
Assume we have a source dataset, $\mathcal{D}_s$, comprising pairs $\{x_s,y_s\}$, where $x_s \in \mathcal{X}_s$ represents images and $y_s \in \mathcal{Y}_s$ denotes the corresponding ground-truth labels. Let $\mathcal{D}_t$ denote the target dataset, consisting solely of a set of images ${x_t}$, where $x_t \in \mathcal{X}_t$. A typical model, $h = g \circ f$, is composed of a feature extractor $f_s: \mathcal{X}_s \rightarrow \mathbb{R}^D$ and a classifier head $g_s: \mathbb{R}^D \rightarrow \mathbb{R}^C$. Here, $D$ represents the channel dimension of the feature vectors, and $C$ is the number of classes. In most network architectures, the classifier head typically consists of a single Multi-Layer Perceptron (MLP) layer with weights $\omega \in \mathbb{R}^{C \times D}$ and biases $\beta \in \mathbb{R}^C$. In the following sections, biases will be omitted for simplicity in our discussion.  Given a source model $h_s$ (e.g., a model trained on the source dataset such as ImageNet~\cite{deng2009imagenet}), the goal of OTTA is to incrementally adapt this source model to the unlabeled target dataset $\mathcal{D}_t$ (such as ImageNet-C~\cite{hendrycks2019robustness}), using each mini-batch of data encountered during inference.

\subsection{Revisiting the Limitations of Entropy Minimization}
Entropy minimization (EM) is a prevalent strategy in current OTTA methodologies, as evidenced by seminal works in the field~\cite{niu2022efficient,niu2023towards,wang2021tent}. In this framework, entropy quantifies the uncertainty in the model's prediction for the target dataset. The entropy of a prediction is mathematically defined as: 
\begin{equation}
    H(p) = -\sum_{c=1}^{C} p_c \log p_c,
\end{equation}
where $p_c$ denotes the predicted probability for class $c$ out of $C$ possible classes. The underlying premise of EM is to minimize this uncertainty, thereby guiding the model to generate more confident predictions on the target dataset.

However, as highlighted in Section~\ref{sec:intro}, we identified two significant drawbacks associated with the application of EM loss function in OTTA: (1) gradient ambiguity near decision boundaries and (2) overconfident predictions for out-of-domain data. 
More specifically, the EM method tends to indiscriminately enhance the logit scores for several top-k classes, which compromises the model’s discriminative accuracy. Moreover, relying exclusively on the EM loss function for OTTA can inadvertently increase the model's confidence in the overconfident yet incorrect predictions, creating an illusion of successful adaptation. Therefore, our findings highlight the need for refined OTTA methodologies that overcome existing limitations of EM approaches to ensure reliable and adaptable model predictions across diverse domains.


\begin{figure}[t]
\centering
\begin{subfigure}[b]{0.244\textwidth}
    \includegraphics[width=\textwidth]{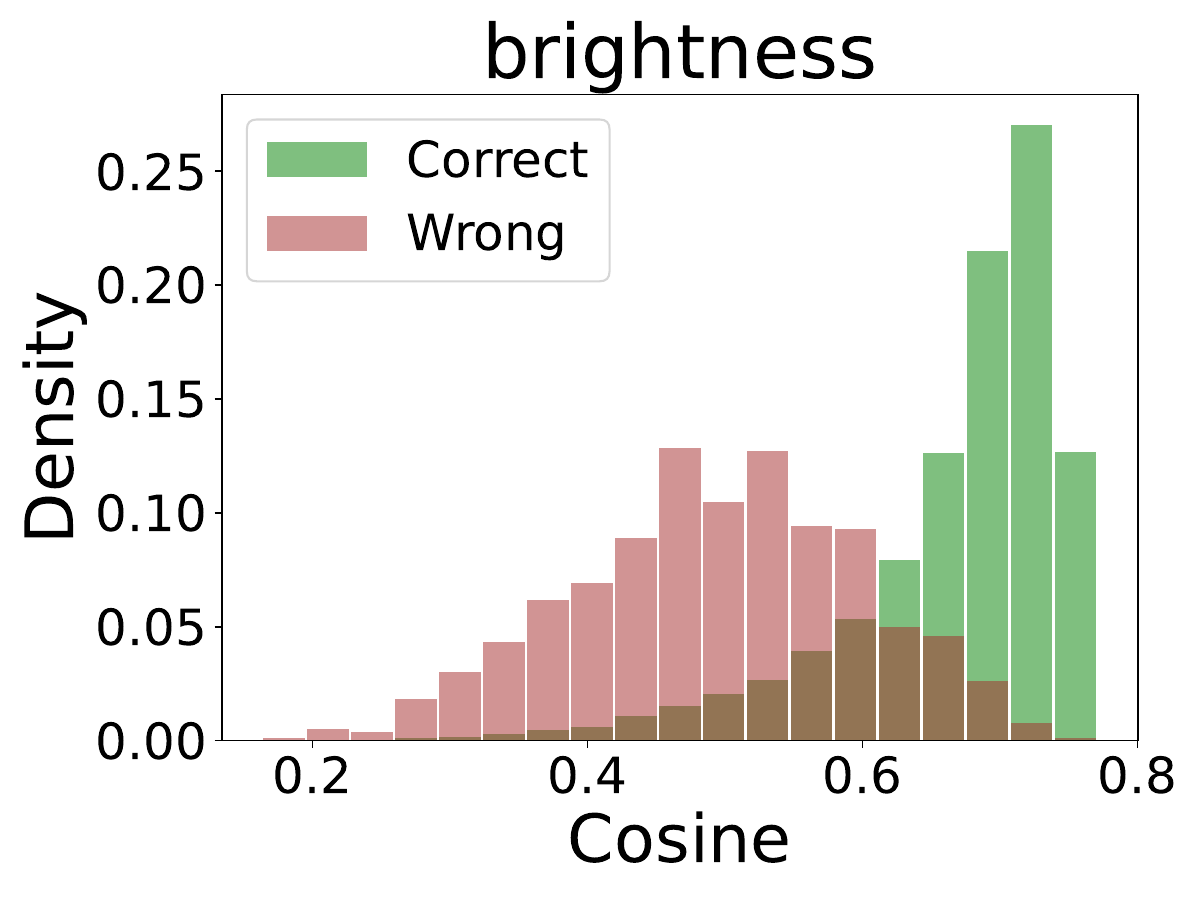}
\end{subfigure}
\hfill 
\begin{subfigure}[b]{0.244\textwidth}
    \includegraphics[width=\textwidth]{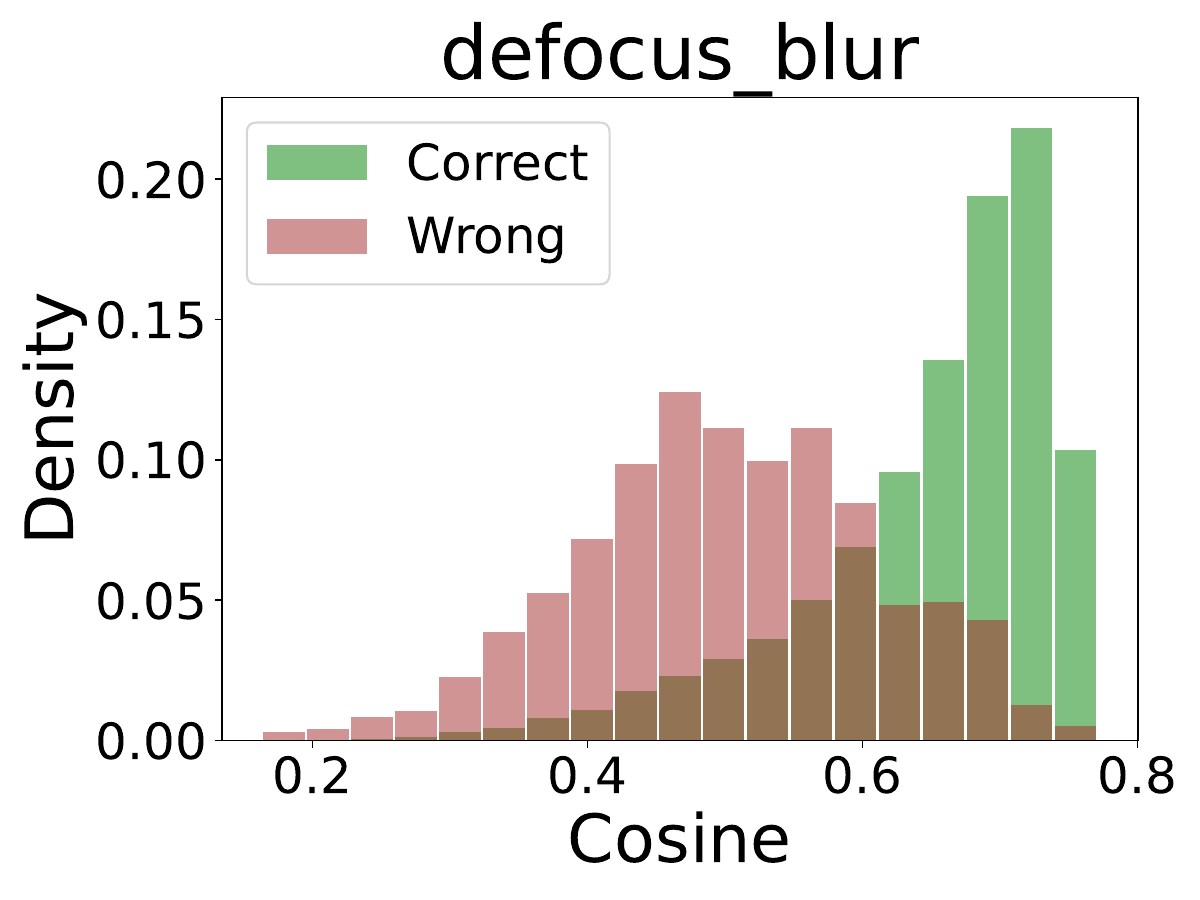}
\end{subfigure}
\hfill 
\begin{subfigure}[b]{0.244\textwidth}
    \includegraphics[width=\textwidth]{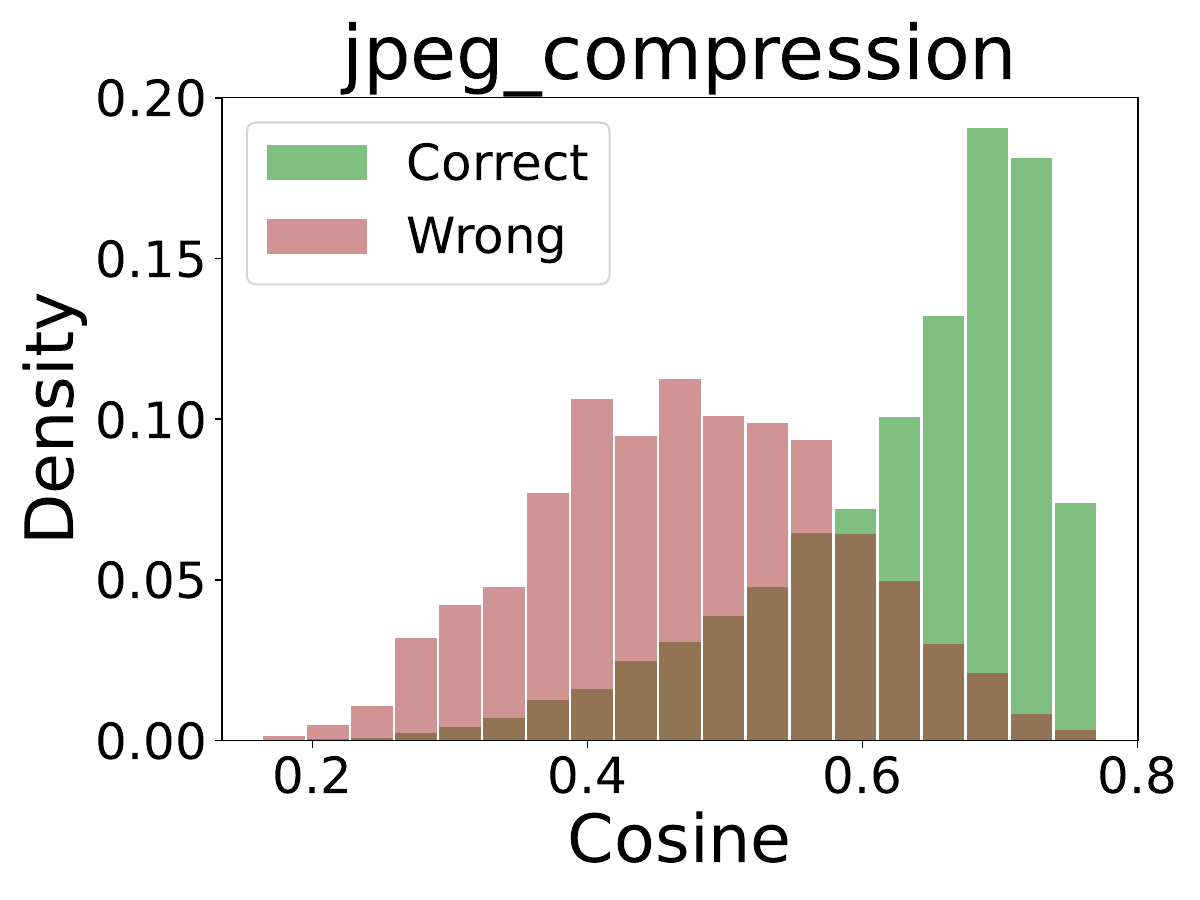}
\end{subfigure}
\hfill 
\begin{subfigure}[b]{0.244\textwidth}
    \includegraphics[width=\textwidth]{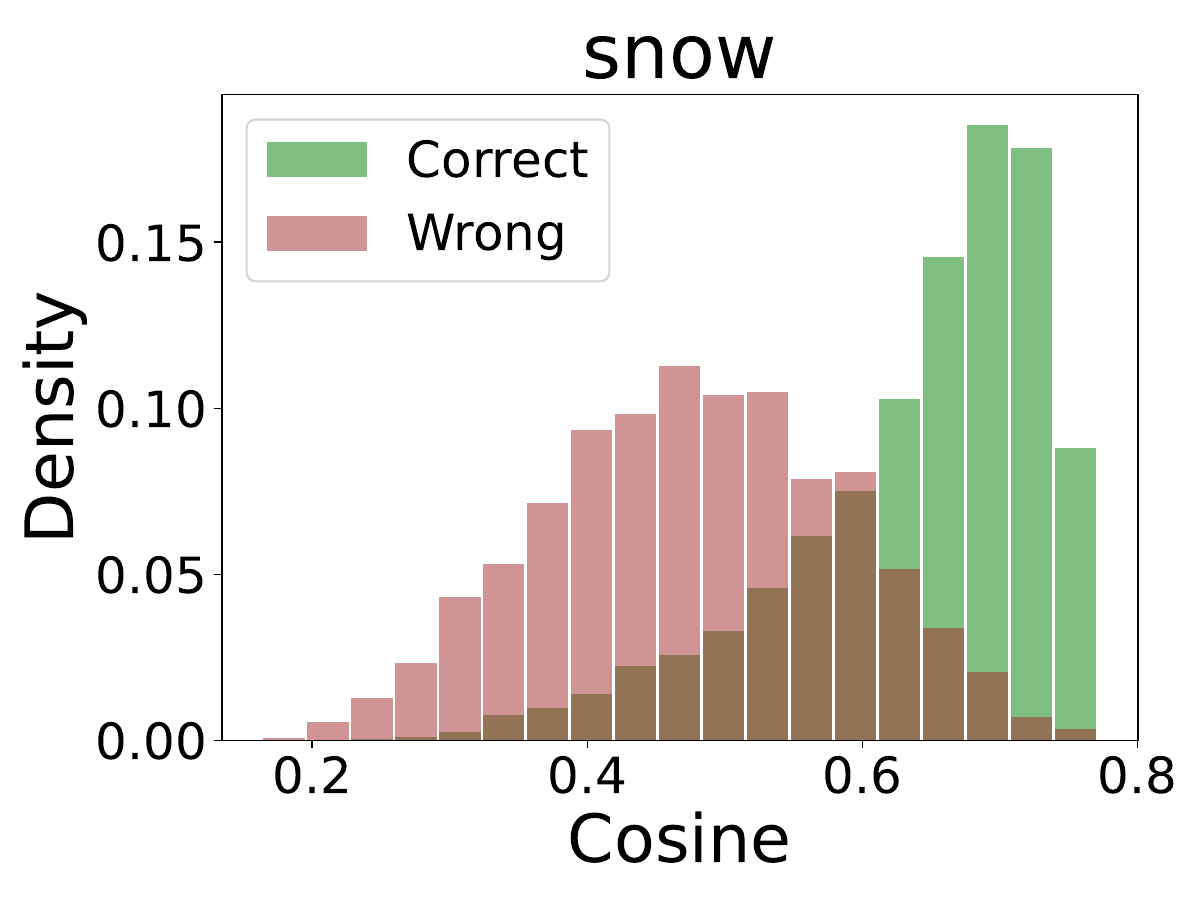}
\end{subfigure}

\caption{The above histograms show cosine similarity distributions of maximally aligned class for correct (green) and incorrect (red) predictions on the CIFAR-10-C dataset, under four different corruption types. A trend is observable where correct predictions are associated with higher cosine similarity scores, while incorrect predictions are more uniformly distributed across the lower range of cosine similarity values. This pattern indicates a strong correlation between higher cosine similarity and prediction accuracy, illustrating the effectiveness of cosine similarity as a measure of correctness amidst data distributional shift, as also suggested in Figure~\ref{fig:motivation} (right).}
\label{fig:max_cosine}
\end{figure}

\subsection{Cosine Alignment Optimization}
In light of the insights gathered, we introduce a novel method that leverages the concept of cosine similarity to enhance OTTA performance. Specifically, our method aims to maximize the cosine similarity between the feature extractor output, $z_i = f(x_i)$, and classifier weight vector for a particular class, $\omega_c$, defined as:
\begin{equation} \label{eqn:cosine}
    cos(\theta_{(\omega_c, z_i)}) = \frac{(\omega_c \cdot z_i)}{\|\omega_c\|\|z_i\|}.
\end{equation}
Here, $i$ index the data points, and $c \in \{ 1, \dots, C\}$ represent a class. Ideally, we would like to maximize the cosine similarity between the feature vector and the true class $c$. However, in the OTTA setting, we do not have access to the true class, making direct optimization of cosine infeasible. Therefore, we instead focus on maximizing the cosine similarity of the maximally aligned, predicted class, computed as:
\begin{equation}
   \hat{c}_i = \max_{c}{\cos(\theta_{\omega_c, z_i})}. 
\end{equation}
In Figure~\ref{fig:max_cosine}, we show that the trends in the cosine similarity of the predicted class correspond with the cosine similarity of the true class depicted in Figure~\ref{fig:motivation}. Specifically, we note that correct predictions correspond to higher cosine similarity scores, whereas incorrect predictions are associated with lower scores within the test dataset (e.g., CIFAR-10-C). This consistent trend confirms the efficacy of cosine similarity of the predicted class as a strong indicator of prediction accuracy, with higher values representing correctness and lower values indicating potential errors.

To improve OTTA performance while addressing the limitations of the entropy minimization loss function, we propose a novel loss function, named CoMM ({\ul \textbf{Co}}sine {\ul \textbf{M}}ax-{\ul \textbf{M}}in). This loss function is designed to enhance the alignment between feature vectors and the weight vector of the predicted class, while also actively penalizing any alignment with the weight vectors of non-predicted classes. The CoMM loss function is mathematically expressed as follows:
\begin{equation} \label{eqn:comm}
    \mathcal{L}_{\text{CoMM}} = -\frac{1}{N}\sum_{i=1}^{N}\log{\frac{{\cos(\theta_{\omega_{\hat{c}_i}, z_i})}}{\sum_j\cos(\theta_{\omega_j, z_i})}}.
\end{equation}
We found that the dual-objective nature of the proposed CoMM loss function plays a pivotal role in significantly enhancing OTTA performance, beyond what is achieved by merely aligning feature vectors with the class weight vector of the predicted class. As detailed in Section~\ref{sec:discussion}, focusing solely on this alignment yields only marginal improvements in OTTA performance. In contrast, our CoMM loss function, which concurrently promotes alignment with the class weight vector of the predicted class and discourages alignment with those of non-predicted classes, consistently delivers superior results across various benchmarks.

It is important to note that our CoMM loss function not only surpasses existing state-of-the-art approaches, including CAFA~\cite{jung2023cafa} and ActMAD~\cite{mirza2023actmad}, in terms of benchmark performances but also unlike these methods, it does not need to access source training data. Thus, our method fully adheres to the assumption of the test-time adaptation problem, where source data is inaccessible while achieving better performance.

\section{Experimental Results}
Following the established practices in the field, the evaluation results of our proposed method are presented in two distinct contexts: (1) robustness against various corruptions and (2) domain adaptation extending beyond mere image corruption scenarios. These evaluations are compared to selected baseline methods, detailed in the following section.

\vspace{5pt}
\noindent \textbf{Baselines}: The primary goal of our experimental evaluation is to compare the effectiveness of our proposed CoMM loss function against the EM loss function for Online Test-Time Adaptation (OTTA). In this context, TENT~\cite{wang2021tent}, distinguished by its use of EM loss function, is selected as the baseline for our studies. To ensure a thorough comparison, we additionally include a variety of recent and impactful OTTA methods as comparison baselines. These methods have been carefully selected to cover a broad spectrum of approaches within the field:
\begin{itemize}
    \item \textbf{Feature Alignment Methods}: Including CFA~\cite{kojima2022robustifying}, CAFA~\cite{jung2023cafa}, and ActMAD~\cite{mirza2023actmad}, these approaches prioritize aligning feature distributions between the source and target domains to mitigate distributional shifts.
    \item \textbf{Pseudo-Prototype Methods}: Such as T3A~\cite{iwasawa2021test}, which adapt the classifier's hypothesis for test data through the utilization of pseudo-prototypes, aiming to recalibrate the model's decision boundaries in light of new data.
    \item \textbf{Transformation Invariance Methods}: Represented by TIPI~\cite{nguyen2023tipi}, these methods enforce the model's invariance to specific transformations during test time, enhancing robustness to variations not encountered during training.
    \item \textbf{Entropy Balancing Methods}: Exemplified by SHOT~\cite{liang2020we}, which focuses on the dual objectives of minimizing prediction entropy to foster decision confidence and maximizing class entropy to ensure even distribution across classes in test data predictions.
\end{itemize}

\begin{table}[t]
\centering
\caption{Top-1 Classification Error (\%) for all corruptions in CIFAR-10C (level 5). Lower is better. All results are for a WRN-40-2 backbone. Source denotes the performance on the corrupted test data without any adaptation. The lowest error rates are highlighted in bold for the best performance.}
\resizebox{\columnwidth}{!}{%
\begin{tabular}{l||ccccccccccccccc||c}
\hline
\multicolumn{1}{c||}{Methods} & Gauss & Shot & Impul & Defcs & Gls  & Mtn  & Zm  & Snw  & Frst & Fg            & Brt & Cnt  & Els  & Px   & Jpg  & Mean \\ \hline
Source                            & 28.8  & 22.9 & 26.2  & 9.5   & 20.6 & 10.6 & 9.3 & 14.2 & 15.3 & 17.5          & 7.6 & 20.9 & 14.7 & 41.3 & 14.7 & 18.3 \\ \hline
SHOT (Offline)                    & 13.4  & 11.6 & 16.3  & 7.3   & 15.9 & 8.2  & 7.1 & 9.4  & 9.4  & 10.2          & 6.3 & 8.3  & 12.8 & 9.8  & 13.6 & 10.6 \\ \hline
TIPI                              & 15.8  & 13.8 & 19.2  & 9.4   & 20.0 & 11.0 & 9.2 & 11.8 & 12.4 & 13.7          & 7.5 & 13.4 & 15.7 & 12.9 & 15.4 & 13.4 \\
T3A                               & 15.7  & 13.9 & 17.8  & 7.9   & 18.2 & 9.0  & 8.2 & 10.9 & 9.7  & 12.6          & 6.1 & 9.2  & 13.4 & 14.2 & 14.4 & 12.1 \\
P-L                               & 15.8  & 14.1 & 17.8  & 7.8   & 18.1 & 8.9  & 8.0 & 10.8 & 9.7  & 12.4          & 6.1 & 9.3  & 13.4 & 14.1 & 14.5 & 12.0 \\
CFA                               & 15.8  & 13.8 & 17.9  & 7.8   & 18.2 & 9.0  & 8.1 & 10.7 & 9.6  & 12.4          & 6.1 & 9.3  & 13.5 & 13.7 & 14.5 & 12.0 \\
SHOT (Online)                     & 14.5  & 12.3 & 17.7  & 7.8   & 17.8 & 8.7  & 7.9 & 10.4 & 9.6  & 12.1          & 6.1 & 9.0  & 13.4 & 11.4 & 14.4 & 11.5 \\
TENT                              & 14.5  & 12.4 & 17.7  & 7.7   & 17.7 & 8.8  & 7.9 & 10.3 & 9.6  & 12.0          & 6.1 & 9.0  & 13.4 & 11.3 & 14.5 & 11.5 \\
ActMADt                           & 13.0  & \textbf{11.2} & 15.1  & 7.4   & 15.9 & 8.3  & 7.1 & 9.5  & 9.3  & 10.6          & \textbf{5.9} & 8.4  & \textbf{12.3} & 9.3  & 13.6 & 10.4 \\ \hline
Ours (CoMM) &
  \textbf{12.9} &
  {11.3} &
  \textbf{14.7} &
  \textbf{7.0} &
  \textbf{15.2} &
  \textbf{8.1} &
  \textbf{7.0} &
  \textbf{8.8} &
  \textbf{8.6} &
  \textbf{8.8} &
  {6.0} &
  \textbf{7.3} &
  {12.4} &
  \textbf{8.9} &
  \textbf{13.0} &
  10.0 \\ \hline
\end{tabular}%
}
\label{tab:cifar10-table}
\end{table} 
\begin{table}[t]
\centering
\caption{Top-1 Classification Error (\%) for all corruptions in CIFAR-100C (level 5). Lower is better. The results were obtained by adapting a WRN-40-2 backbone, trained on CIFAR100, to CIFAR-100C.}
\label{tab:cifar100-table}
\resizebox{\columnwidth}{!}{%
\begin{tabular}{l||ccccccccccccccc||c}
\hline
\multicolumn{1}{c||}{Methods} & Gauss & Shot & Impul & Defcs         & Gls  & Mtn  & Zm   & Snw  & Frst & Fg            & Brt  & Cnt  & Els  & Px   & Jpg  & Mean \\ \hline
Source                           & 65.7  & 60.1 & 59.1  & 32.0          & 51.0 & 33.6 & 32.4 & 41.4 & 45.2 & 51.4          & 31.6 & 55.5 & 40.3 & 59.7 & 42.4 & 46.8 \\ \hline
SHOT (Offline)                   & 37.2  & 36.2 & 36.7  & 27.5          & 38.2 & 28.5 & 27.8 & 31.8 & 32.0 & 33.4          & 25.8 & 29.6 & 34.5 & 29.8 & 37.2 & 32.4 \\ \hline
T3A                              & 42.4  & 41.8 & 42.5  & 29.7          & 44.3 & 30.5 & 29.5 & 35.9 & 34.5 & 42.1          & 26.8 & 32.8 & 38.0 & 35.9 & 40.7 & 36.5 \\
TIPI                             & 40.9  & 39.7 & 42.2  & 30.6          & 42.9 & 31.5 & 30.5 & 35.8 & 35.8 & 40.2          & 28.4 & 34.7 & 38.3 & 33.4 & 41.4 & 36.4 \\
P-L                              & 41.3  & 40.5 & 42.5  & 29.6          & 43.1 & 30.3 & 29.4 & 35.8 & 34.3 & 41.7          & 26.7 & 32.4 & 37.8 & 33.5 & 40.8 & 36.0 \\
CFA                              & 40.4  & 39.3 & 42.1  & 29.4          & 42.3 & 30.2 & 29.2 & 35.1 & 34.1 & 39.8          & 26.7 & 32.1 & 37.6 & 32.8 & 40.6 & 35.5 \\
SHOT (Online)                    & 39.7  & 38.9 & 42.1  & {29.0} & 41.9 & 30.2 & 29.3 & 34.8 & 34.2 & {39.7} & 26.7 & 32.2 & 37.2 & 32.5 & 40.4 & 35.3 \\
TENT                             & 39.9  & 39.1 & 42.2  & {29.0} & 42.0 & 30.2 & 29.3 & 34.9 & 34.2 & {39.7} & 26.7 & 32.3 & 37.4 & 32.4 & 40.4 & 35.3 \\
ActMADt &
  {39.6} &
  {38.4} &
  {39.5} &
  29.1 &
  {41.5} &
  {30.0} &
  {29.1} &
  {34.0} &
  {33.2} &
  40.2 &
  {26.4} &
  {31.5} &
  {36.4} &
  {31.4} &
  {38.9} &
  34.6 \\ \hline
Ours (CoMM) &
  \textbf{38.1} &
  \textbf{36.6} &
  \textbf{36.4} &
  \textbf{27.4} &
  \textbf{39.2} &
  \textbf{28.7} &
  \textbf{27.6} &
  \textbf{31.9} &
  \textbf{32.1} &
  \textbf{33.2} &
  \textbf{25.8} &
  \textbf{28.0} &
  \textbf{35.2} &
  \textbf{29.7} &
  \textbf{38.1} &
  \textbf{32.5} \\ \hline
\end{tabular}%
}
\end{table}

\noindent \textbf{Implementation Details}:
For evaluating robustness to corruptions, we adopt the implementation approach outlined by Mirza~\etal~\cite{mirza2023actmad} by utilizing the WRN-$40\times2$ model on the CIFAR-10-C and CIFAR-100-C datasets. In addition, our experiments on the ImageNet-C dataset utilized both ResNet18 and ResNet50 models to ensure a comprehensive analysis. For domain adaptation experiments on the OfficeHome~\cite{venkateswara2017deep} and DomainNet~\cite{peng2019moment} datasets, the ResNet50 model was chosen. We utilized publicly available pre-trained models for our experiments: the AugMix pre-trained WRN-$40\times2$ model~\cite{hendrycks2020augmix} and the ImageNet-1k pre-trained ResNet18/50 models. 

Similarly, following the implementation protocol established by ActMAD~\cite{mirza2023actmad}, we set the batch size to 128 for OTTA experiments on robustness to corruption across CIFAR10-C, CIFAR100-C, and ImageNet-C datasets (specifically with ResNet18). This batch size is consistently applied to our domain adaptation experiments as well. For ImageNet-C experiments employing ResNet50, we adjust the batch size to 64, aligning with the methodologies of prior works~\cite{jung2023cafa, nguyen2023tipi}. Regarding learning rates, we adopt a strategy tailored to each dataset: 0.005 for CIFAR-10-C, CIFAR-100-C and ImageNet-C, and 0.00025 for domain adaptation tasks. We utilized Stochastic Gradient Descent (SGD) as the optimizer for all our experimental setups. Following the footsteps of existing approaches, our experimental setup focuses on optimizing only the learnable affine parameters within the normalization layers, a decision driven by efficiency considerations~\cite{wang2021tent}. Empirically, we found that our approach remains effective even when applied to optimizing the entire feature extractor network of a model. 

\useunder{\uline}{\ul}{}
\begin{table}[t]
\centering
\caption{Top-1 Classification Error (\%) for all corruptions in ImageNet-C (level 5). Lower is better. All results are for a ResNet-18 and ResNet-50 network pre-trained on the clean ImageNet-1k train set. Source denotes its performance on the corrupted test data without any adaptation. Results marked with † were obtained from our implementation. The lowest error rates are highlighted in \textbf{bold} for the best performance, and the second-best results are {\ul underlined} for comparative emphasis.}
\label{tab:imagenet-table}
\resizebox{\textwidth}{!}{%
\begin{tabular}{lcccccccccccccccc}
\hline
\multicolumn{1}{c|}{{Methods}} &
  {Gauss} &
  {Shot} &
  {Impul} &
  {Defcs} &
  {Gls} &
  {Mtn} &
  {Zm} &
  {Snw} &
  {Frst} &
  {Fg} &
  {Brt} &
  {Cnt} &
  {Els} &
  {Px} &
  \multicolumn{1}{c|}{{Jpg}} &
  {Mean} \\ \hline
\multicolumn{17}{c}{ResNet-18}                                                                                                                                            \\ \hline
\multicolumn{1}{l|}{Source}          & 98.4 & 97.7 & 98.4 & 90.6 & 92.5 & 89.8 & 81.8 & 89.5 & 85.0 & 86.3 & 51.1 & 97.2 & 85.3 & 76.9 & \multicolumn{1}{c|}{71.7} & 86.2 \\ \hline
\multicolumn{1}{l|}{SHOT (Offline)}  & 73.8 & 70.5 & 72.2 & 79.2 & 80.6 & 58.5 & 54.0 & 53.6 & 63.0 & 47.3 & 39.2 & 97.7 & 48.7 & 46.1 & \multicolumn{1}{c|}{53.0} & 62.5 \\ \hline
\multicolumn{1}{l|}{T3A}             & 85.5 & 84.0 & 85.0 & 86.6 & 85.9 & 76.1 & 65.4 & 70.3 & 71.0 & 58.7 & 41.3 & 86.8 & 60.5 & 54.4 & \multicolumn{1}{c|}{61.0} & 71.5 \\
\multicolumn{1}{l|}{SHOT (Online)}   & 83.9 & 82.3 & 83.7 & 83.9 & 83.8 & 72.6 & 61.9 & 65.7 & 68.6 & 54.8 & 39.4 & 85.9 & 58.1 & 53.1 & \multicolumn{1}{c|}{62.3} & 69.3 \\
\multicolumn{1}{l|}{P-L}             & 82.0 & 79.7 & 81.5 & 84.2 & 83.0 & 71.0 & 60.7 & 65.4 & 68.6 & 52.9 & 41.7 & 82.6 & 55.5 & 51.1 & \multicolumn{1}{c|}{55.7} & 67.7 \\
\multicolumn{1}{l|}{TENT}            & 80.8 & 78.6 & 80.4 & 82.5 & 82.5 & 72.1 & 60.5 & 63.7 & 66.7 & 52.1 & {\ul 39.2} & 84.2 & 55.5 & 50.8 & \multicolumn{1}{c|}{58.2} & 67.2 \\
\multicolumn{1}{l|}{CFA}             & 78.2 & 76.4 & 78.2 & 81.9 & 80.4 & 69.6 & 60.1 & 63.4 & 67.6 & 52.0 & 41.5 & {\ul 79.5} & 54.3 & 50.2 & \multicolumn{1}{c|}{55.1} & 65.9 \\
\multicolumn{1}{l|}{ActMAD}          & {\ul 76.3} & 77.4 & 77.4 & \textbf{76.1} & \textbf{75.4} & 72.0 & 62.8 & 66.6 & 65.8 & 55.8 & 40.9 & \textbf{78.8} & 55.7 & 51.4 & \multicolumn{1}{c|}{57.6} & 66.0 \\
\multicolumn{1}{l|}{TIPI$^{\dagger}$}            & \textbf{76.1} & 74.5 & \textbf{76.1} & {\ul 79.0} & 77.6 & {\ul 67.6} & {\ul 58.7} & {\ul 62.5} & \textbf{65.4} & {\ul 50.6} & 40.2 & {\ul 79.5} & {\ul 52.6} & {\ul 48.4} & \multicolumn{1}{c|}{{\ul 53.0}} & {\ul 64.1} \\ \hline
\multicolumn{1}{l|}{Ours (CoMM)}            & {\ul 76.3} & \textbf{73.6} & {\ul 76.3} & \textbf{76.1} & {\ul 77.3} & \textbf{59.9} & \textbf{53.7} & \textbf{56.5} & {\ul 65.6} & \textbf{47.1} & \textbf{38.4} & 87.7 & \textbf{49.3} & \textbf{46.0} & \multicolumn{1}{c|}{\textbf{51.1}} & \textbf{62.3} \\ \hline
\multicolumn{17}{c}{ResNet-50}                                                                                                                                            \\ \hline
\multicolumn{1}{l|}{Source}          & 97.8 & 97.1 & 98.2 & 82.1 & 90.2 & 85.2 & 77.5 & 83.1 & 76.7 & 75.6 & 41.1 & 94.6 & 83.1 & 79.4 & \multicolumn{1}{c|}{68.4} & 82.0 \\
\multicolumn{1}{l|}{TENT}            & 71.6 & 69.8 & 69.9 & 71.8 & 72.7 & 58.6 & 50.5 & 52.9 & 58.7 & 42.5 & {\ul 32.6} & 74.9 & 45.2 & 41.5 & \multicolumn{1}{c|}{47.7} & 57.4 \\
\multicolumn{1}{l|}{TIPI$^{\dagger}$}& {\ul 69.0} & 68.1 & \textbf{67.8} & {\ul 70.6} & {\ul 70.6} & 57.4 & {\ul 49.5} & 52.5 & 57.3 & {\ul 41.8} &  \textbf{32.0} & {66.6} & 44.0 & {\ul 40.1} & \multicolumn{1}{c|}{\ul 45.9} & 55.5 \\
\multicolumn{1}{l|}{CAFA}            & 69.6 &  \textbf{67.3} & {\ul 68.0} & 71.1 & 70.9 & {\ul 56.1} & 50.0 & {\ul 50.8} & {\ul 56.8} & 41.9 & {33.2} & \textbf{61.3} & {\ul 43.8} & 40.9 & \multicolumn{1}{c|}{47.0} & {\ul 55.2} \\ \hline
\multicolumn{1}{l|}{Ours (CoMM)}            & \textbf{68.3} & {\ul 68.0} & {68.5} & \textbf{68.2} & \textbf{69.0} & \textbf{53.9} & \textbf{48.1} & \textbf{49.4} & \textbf{55.8} & \textbf{40.5} & \textbf{32.0} & {\ul 61.6} & \textbf{42.3} & \textbf{39.5} & \multicolumn{1}{c|}{\textbf{45.1}} & \textbf{53.9} \\ \hline
\end{tabular}%
}
\end{table}
\subsection{Robustness to Corruptions}
\noindent \textbf{Datasets}: Our evaluation of method robustness against data corruptions utilizes the CIFAR10-C, CIFAR100-C, and ImageNet-C datasets. The base CIFAR10 and CIFAR100 datasets each consist of $50,000$ training samples and 10,000 test samples, across $10$ and $100$ classes, respectively. ImageNet, significantly larger, comprises $1.2$ million training samples and $50,000$ validation samples, spanning 1,000 object categories. The CIFAR10-C and CIFAR100-C datasets introduce 15 distinct types of corruptions to the test sets of their respective base datasets. Similarly, the ImageNet-C dataset applies the same $15$ types of corruptions, but to the validation set of ImageNet, providing a comprehensive framework for assessing the resilience of methods in the face of diverse and challenging data distortions. 

\noindent \textbf{Quantitative evaluation and comparisons}: Tables~\ref{tab:cifar10-table} and \ref{tab:cifar100-table} present a performance comparison between our CoMM loss function and the state-of-the-art methods on CIFAR-10-C and CIFAR-100-C datasets, respectively. Notably, our method achieves consistently higher performance on both datasets, with a substantial margin of improvement observed in the CIFAR-100-C dataset. This improved performance is not only consistent but also significant when directly compared to TENT~\cite{wang2021tent} that relies on entropy minimization, affirming the robustness of our approach against the entropy minimization technique. Furthermore, our method demonstrates competitive results in line with offline methods such as SHOT~\cite{liang2020we}, without requiring pre-computed classwise-centroids in the target domain. The ability to bypass such preparatory steps without compromising on accuracy underscores the practicality and adaptability of our method, making it a robust approach for scenarios where real-time adaptation is critical and computational efficiency is desired.


Additionally, our method also demonstrates its superiority within the challenging ImageNet-C benchmark. As shown in Table~\ref{tab:imagenet-table}, our method outperforms existing works, including TENT and CAFA~\cite{jung2023cafa}, across a spectrum of corruptions. Particularly, the results with ResNet-18 and ResNet-50 models underscore the robustness of our method, achieving a notable top-1 classification error rate of $62.3\%$ and $53.6\%$ on ResNet18 and ResNet50, respectively. This performance is not just an improvement over traditional entropy minimization approaches but also able to achieve performance on par with (in some cases superior to) offline methods that require additional pre-processing steps. The effectiveness of our method in handling the diverse and complex corruptions in ImageNet-C, without the need for prior domain knowledge, highlights its potential as a robust and reliable tool for real-world applications where models must contend with unpredictable and varied data shifts.

\useunder{\uline}{\ul}{}
\begin{table}[t]
\centering
\caption{Classification error rates (\%) across domain adaptation tasks on the OfficeHome dataset. Results marked with $\dagger$ were obtained from our implementation, while others were sourced from~\cite{jung2023cafa}. The best performance (lowest error rates) are highlighted in \textbf{bold}, and the second-best results are {\ul underlined} for comparative emphasis.}
\label{tab:office-home}
\resizebox{\textwidth}{!}{%
\begin{tabular}{l|ccc|ccc|ccc|ccc|c}
\hline
\multicolumn{1}{l|}{Method}        & Ar→Cl          & Ar→Pr          & Ar→Re          & Cl→Ar          & Cl→Pr          & Cl→Re          & Pr→Ar          & Pr→Cl          & Pr→Re          & Re→Ar          & Re→Cl          & \multicolumn{1}{c|}{Re→Pr}          & Average        \\ \hline
\multicolumn{1}{l|}{Source}        & 67.33          & 46.68          & 36.81          & 68.52          & 54.22          & 53.91          & 66.25          & 71.25          & 40.76          & 45.16          & 65.02          & \multicolumn{1}{c|}{29.65}          & 53.80          \\
\multicolumn{1}{l|}{Source$^{\dagger}$} & 67.22          & 45.91          & 35.65          & 67.31          & 53.22          & 53.01          & 65.05          & 72.25          & 39.11          & 44.91          & 64.99          & \multicolumn{1}{c|}{29.69}          & 53.19          \\
\multicolumn{1}{l|}{PL}            & 63.18          & 46.25          & 35.87          & 65.68          & 56.14          & 53.16          & 61.80          & 68.34          & 38.51          & 45.45          & 60.53          & \multicolumn{1}{c|}{29.62}          & 52.04          \\
\multicolumn{1}{l|}{TENT}         & 61.47          & 44.33          & 34.82          & 62.75          & 52.22          & 49.16          & 61.60          & 66.19          & 36.26          & 44.66          & 58.08          & \multicolumn{1}{c|}{28.14}          & 49.97          \\
\multicolumn{1}{l|}{T3A}           & 62.29          & \textbf{41.41} & 34.11          & 64.65          & 51.05          & 48.91          & 60.61          & 66.96          & {\ul 35.69}    & 45.69          & 59.22          & \multicolumn{1}{c|}{{\ul 27.64}}    & 49.85          \\
\multicolumn{1}{l|}{AdaContrast}   & 61.97          & {\ul 41.95}    & 34.59          & 62.46          & 50.96          & 49.71          & {\ul 59.21}    & 65.27          & 36.93          & 46.11          & 56.13          & \multicolumn{1}{c|}{28.23}          & 49.46          \\
\multicolumn{1}{l|}{EATA}         & 62.86          & 43.64          & 34.43          & 63.37          & {\ul 50.91}    & 48.70          & 60.77          & 65.91          & 35.99          & 43.30          & 56.40          & \multicolumn{1}{c|}{27.84}          & 49.43          \\
\multicolumn{1}{l|}{CAFA}          & \textbf{59.73} & 42.64          & {\ul 34.01}    & {\ul 61.39}    & 51.23          & {\ul 47.69}    & 60.28          & \textbf{63.92} & 35.87          & {\ul 42.89}    & \textbf{54.91}    & \multicolumn{1}{c|}{27.84}          & 48.53          \\ \hline
\multicolumn{1}{l|}{Ours (CoMM)}          & {\ul 59.79}    & 42.05          & \textbf{33.13} & \textbf{60.98} & \textbf{49.69} & \textbf{46.49} & \textbf{58.22} & {\ul 64.77}     & \textbf{35.17} & \textbf{40.40} & {\ul 55.11} & \multicolumn{1}{c|}{\textbf{26.47}} & \textbf{47.69} \\ \hline
\end{tabular}%
}
\end{table}

\begin{table}[t]
    \centering
    \begin{minipage}{0.54\textwidth}
        \centering
        \caption{Classification Error Rates (\%) on the DomainNet Dataset. Each column specifies a different source domain, with the reported error rates representing the average performance across all target domains.}
        \label{tab:domainNet}
        \resizebox{\textwidth}{!}{%
        \begin{tabular}{l|cccccc|c}
        \hline
        Method & Clip. & Info. & Pain. & Quic. & Real  & Sket. & Avg.  \\ \hline
        Source & 76.53 & 75.38 & 74.29 & 96.69 & 73.16 & 75.18 & 78.54 \\
        BN     & 76.14 & 79.25 & 72.90 & 93.15 & 74.17 & 68.80 & 77.40 \\
        PL     & 75.36 & 78.02 & 72.47 & 93.01 & 73.21 & 68.17 & 76.71 \\
        TENT  & 84.59 & 75.66 & 71.81 & 92.86 & 71.60 & 67.85 & 77.40 \\
        T3A    & 74.10 & 76.50 & 71.68 & 92.89 & 73.50 & 66.64 & 75.89 \\
        EATA  & 73.88 & 75.26 & 71.16 & 92.37 & 70.69 & 66.86 & 75.04 \\
        CAFA   & 73.17 & 74.69 & 71.05 & 92.49 & 69.96 & 66.53 & 74.65 \\ \hline
        CoMM & \textbf{73.02} & \textbf{73.52} & \textbf{70.85} & \textbf{87.70} & \textbf{68.66} & \textbf{67.24} & \textbf{73.50} \\ \hline
        \end{tabular}%
        }
    \end{minipage}
    \hfill
    \begin{minipage}{0.42\textwidth}
        \centering
        \caption{Ablation study comparing the performance of two cosine alignment criteria variants, namely $\mathcal{L}_\text{CoM}$ and $\mathcal{L}_\text{CoMM}$, against the entropy minimization method $\mathcal{L}_\text{EM}$ across three corruption benchmarks: CIFAR-10-C (CF10), CIFAR-100-C (CF100), and ImageNet-C (ImN). } 
        \label{tab:ablation}
        \resizebox{0.9\textwidth}{!}{%
        \begin{tabular}{l@{\hspace{10pt}}|@{\hspace{10pt}}c@{\hspace{10pt}}c@{\hspace{10pt}}c@{\hspace{10pt}}}
        \hline
        \multirow{2}{*}{Methods} & \multicolumn{3}{c}{Classification Error (\%)} \\
                                 & CF10    & CF100    & ImN   \\ \hline
        Source &18.27 &46.76 & 86.15 \\ \hline 
        $\mathcal{L}_\text{EM}$                   & 11.57         & 35.31          & 67.22        \\ 
        $\mathcal{L}_\text{CoM}$                   & 11.10         & 35.28          & 67.77        \\
        $\mathcal{L}_\text{CoMM}$                  & 10.14         & 32.75          & 62.33        \\ \hline
        \end{tabular}%
        }        
    \end{minipage}
    
\end{table}


\subsection{Domain Adaptation beyond Image Corruption Shifts}
\noindent \textbf{Datasets}: For our domain adaptation experiments, we employ the Office-Home and DomainNet datasets, which are benchmarks in the domain adaptation field. The Office-Home dataset is composed of approximately $15,500$ images across $65$ categories of everyday objects, distributed among four unique domains: Artistic images (Ar), Clipart images (Cl), Product images (Pr), and Real-world images (Re). The DomainNet dataset, considerably larger, consists around $600,000$ images spanning $345$ categories over six domains: clipart, infograph, painting, quickdraw, real, and sketch.
\vspace{5pt}

\noindent \textbf{Quantitative Evaluation and Comparisons}: As shown in Table~\ref{tab:office-home}, our method outperforms the baseline method by a margin of $5.5\%$ and advances beyond TENT~\cite{wang2021tent} by $2.3\%$ on average across all domain shifts in the Office-Home dataset. It also attains performance improvement over EATA~\cite{niu2022efficient}, which employs a selective entropy minimization strategy. Moreover, as illustrated in Table~\ref{tab:domainNet}, our approach reduces the overall average error rate to $73.50\%$, demonstrating an improvement over the baseline and state-of-the-art methods including TENT by a respectable margin on the DomainNet benchmark. These results validate the robust adaptability of our method and suggest its potential to become a new benchmark for domain adaptation tasks.
\vspace{-10pt}

\section{Discussion} \label{sec:discussion}
\vspace{-5pt}
In the following sections, we conduct ablation studies to objectively compare the performance of the proposed CoMM loss function with the entropy minimization (EM) loss function and another loss function that solely aligns feature vectors with the weight vector of the predicted class. For analysis purposes, we designed another loss function named CoM ({\ul\textbf{Co}}sine {\ul \textbf{M}}ax) that aligns solely with the weight vector of the predicted class, which is defined as: 
\begin{equation} \label{eqn:com}
\mathcal{L}_{\text{CoM}} = \frac{1}{N}\sum_{i=1}^{N}\arccos \left ({\max_{c}{\cos(\theta_{\omega_c, z_i})}}\right ),
\end{equation}
where $N$ is the number of samples in a mini-batch and the $\arccos (\cdot)$ function transforms the objective from a maximizing cosine similarity to a minimization problem. Following that, we extend our analysis to compare the gradients induced by the EM, hard pseudo-labeling (PL), and our cosine similarity-based loss functions (CoM and CoMM) and analyze their performance across various batch sizes. These analyses offer insights into how each method contributes to the OTTA performance. \\ 

\begin{figure}[t]
\centering
\begin{minipage}{0.45\textwidth}
\centering
    \includegraphics[width=0.9\textwidth]{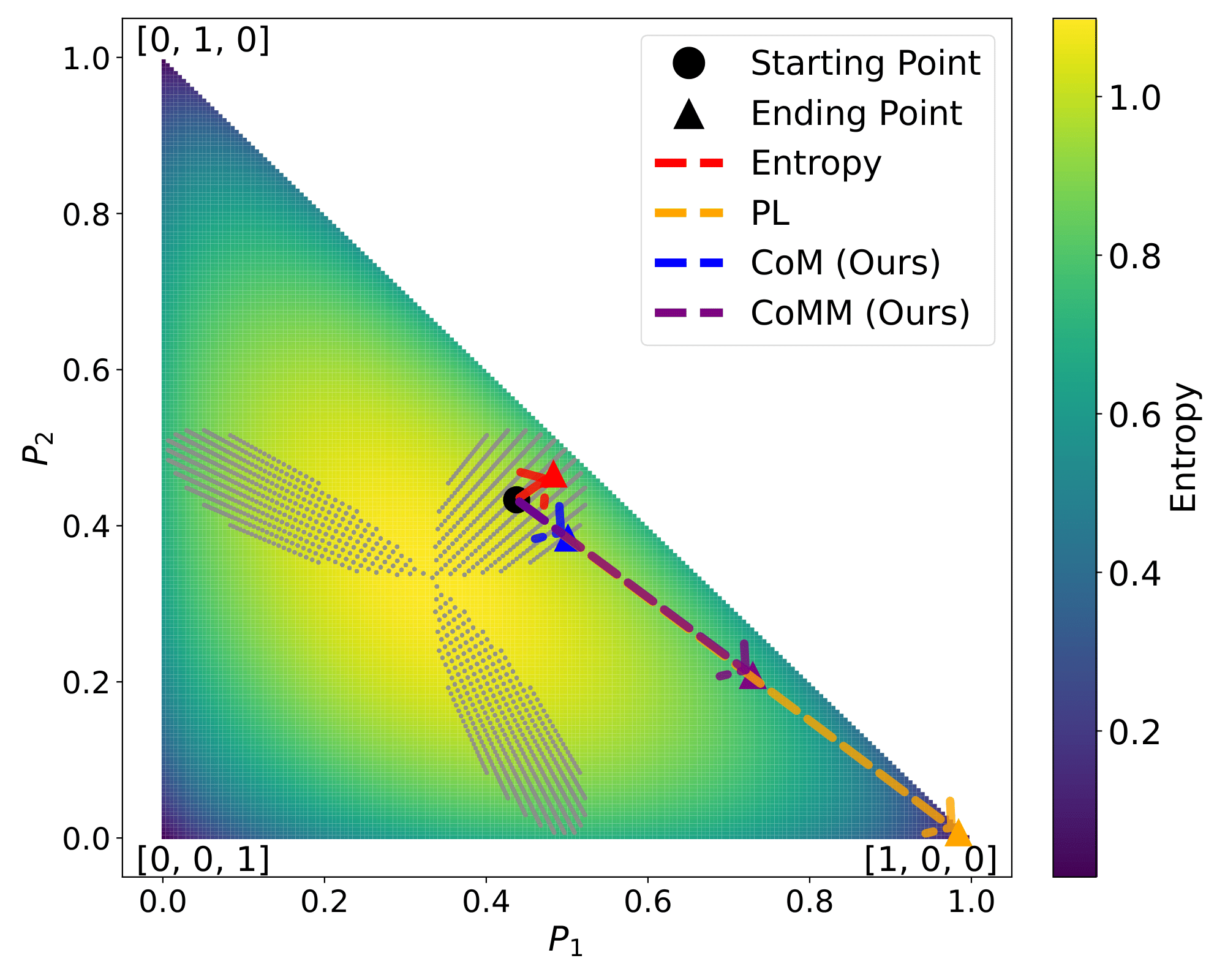}
    \caption{Visualization of optimization trajectories on the probability simplex for various online test-time adaptation methods, namely entropy minimization (EM), hard pseudo label (PL), CoM and CoMM loss functions. The color gradient indicates entropy levels, with darker regions corresponding to lower entropy.}
    \label{fig:simplex_gradient}
\end{minipage}
\hfill
\begin{minipage}{0.51\textwidth}
\centering
    \includegraphics[width=\textwidth]{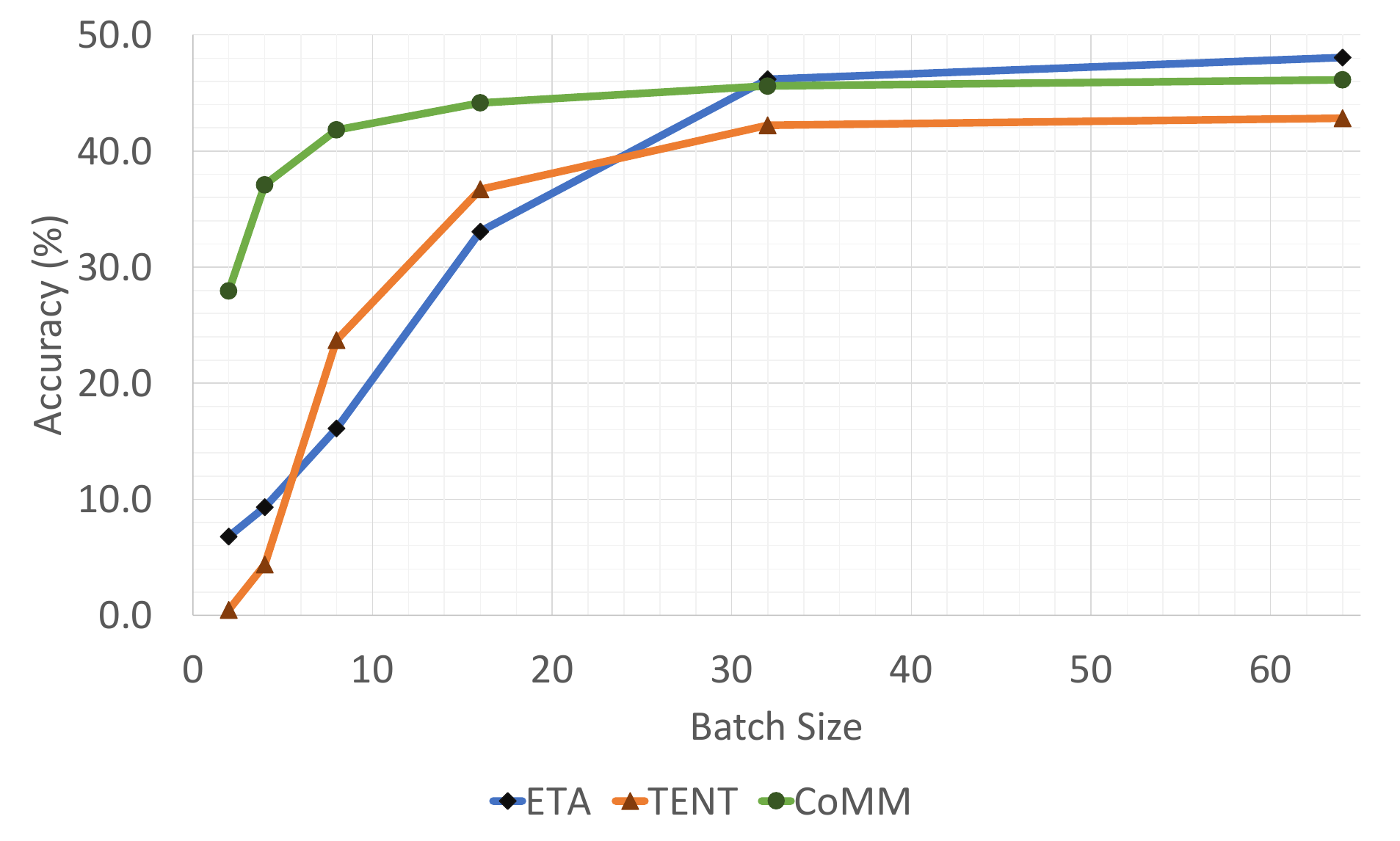}
    \caption{Performance comparison of OTTA methods against varying batch sizes, showcasing our CoMM approach alongside entropy minimization methods such as TENT and ETA. While TENT and ETA experience significant performance drops with batch sizes under $32$, CoMM maintains a higher level of accuracy consistently across batch sizes.}
    \label{fig:batch_dependency}
\end{minipage}
\vspace{-5pt}
\end{figure}

\noindent \textbf{Ablation Study}: 
As illustrated in Table~\ref{tab:ablation}, the CoM loss function, which focuses solely on aligning with the predicted class, can improve model adaptation on novel datasets and surpasses the performance of the source model. However, when compared against the CoMM loss function that incorporates the dual objective of promoting alignment with the predicted class while penalizing alignment with other classes, CoMM consistently outperforms CoM by a large margin across three different benchmarks. This validates the significance of enforcing the dual-objective strategy inherent in our proposed CoMM loss function for enhanced OTTA. 

Furthermore, Table~\ref{tab:ablation} also shows that our proposed CoMM surpasses the performance of the EM loss function, achieving reductions in classification error rates of $1.43\%$, $2.56\%$ and $4.89\%$ for the CIFAR-10-C, CIFAR-100-C and ImageNet-C datasets, respectively. These results confirm the effectiveness of CoMM in enhancing online test-time adaptation across diverse settings.

\vspace{5pt}

\noindent \textbf{Comparative Analysis of Entropy Minimization, Hard Pseudo Label, and CoMM}: 
Figure~\ref{fig:simplex_gradient} analytically compares EM, PL, and our proposed cosine-alignment-based loss functions CoM and CoMM, in their effectiveness in enhancing model certainty and reducing prediction entropy. The EM loss function (red line) aims to reduce uncertainty by guiding predictions towards class-specific simplex vertices, but fails to ensure accurate class convergence, particularly in cases of high initial uncertainty (e.g., low-margin samples in gray areas). Meanwhile, PL optimizes towards the most probable class, proving effective when initial predictions are accurate but potentially misleading otherwise.

CoM (blue line) illustrates a trajectory towards improved accuracy but may struggle to reduce uncertainty in cases of initial predictions with low confidence. Given that OTTA typically applies a single optimization step per sample or batch, the inability of CoM to address these situations results in subpar performance, especially for these samples with low confidence. On the other hand, CoMM (purple line) introduces a dual-objective approach, enhancing alignment with the predicted class while reducing it with others, demonstrating a clear optimization trajectory towards the correct class vertex, significantly reducing the chances of ambiguous predictions and showcasing its comprehensive effectiveness in the OTTA context.

Distinct from PL, which is susceptible to quick convergence towards incorrect predictions due to overly confident initial outputs, CoMM utilizes a regularized optimization to generate less aggressive gradients. This approach notably reduces the risk of overfitting, especially in scenarios involving smaller batch sizes where such risk is elevated. For instance, with a batch size of $8$ on the ImageNet-C benchmark, CoMM achieves an accuracy of $41.8\%$, surpassing the $12.7\%$ of PL and $23.7\%$ of EM, thus demonstrating the effective adaptation and robustness of our proposed CoMM loss function for OTTA. 
\vspace{5pt}

\noindent \textbf{Batch Dependency}: 
This section discusses the impact of different batch sizes on the performance of our proposed method. As pointed out by Zhao~\etal~\cite{zhao2023ttab}, existing OTTA techniques, including entropy minimization methods such as TENT~\cite{wang2021tent}, exhibit a ``batch dependency'' issue, where their performance deteriorates with smaller batch sizes or in the presence of individual samples. Such conditions can render batch-based methods impractical. As depicted in Figure~\ref{fig:batch_dependency}, TENT's performance deteriorates with smaller batches (e.g., less than 32). A similar trend is also observed in strategies like ETA~\cite{niu2022efficient}, which, despite improvements for larger batches by excluding high entropy samples, also experiences a reduction in performance with decreased batch sizes.

In contrast, adapting a model with our proposed CoMM loss function demonstrates a more consistent and stable performance relative to these methods, even when constrained to small batches, \textit{without any specific tailoring to address batch dependency concerns}. This resilience can be attributed to the inherent design of our CoMM loss function, which optimizes the feature vector's alignment with the class weight vector on a per-sample basis rather than relying on batch-level statistics. However, similar to other approaches, our method also suffers from performance degradation when the batch size is smaller than 8. While this falls outside the scope of the current study, it presents an avenue for future research to enhance OTTA robustness under extreme batch size reductions.

\vspace{-10pt}
\section{Conclusions}


In conclusion, this study highlights the limitations of the entropy minimization loss function in Online Test-Time Adaptation (OTTA), especially its inadequacy in handling low-margin samples and incorrect yet high-confidence predictions. To address these challenges, we have introduced a novel feature-weight cosine alignment optimization strategy, underpinned by a dual-objective loss function. The proposed approach significantly improves prediction accuracy and enhances adaptability to novel domains, presenting a more effective method of alignment than previously employed entropy-based techniques.

Critically, we found that cosine similarity between feature vectors and class weight vectors is a more reliable measure of prediction correctness compared to certainty measured via entropy. Moreover, our method prioritizes targeted alignment through cosine similarity, as it directly aligns feature vectors with class weight vectors, offering a precise and discriminative signal for adaptation. Our extensive empirical results, spanning CIFAR and ImageNet benchmarks and domain adaptation scenarios with the Office-Home and DomainNet datasets, underline the superior performance of our approach. Our method consistently surpasses state-of-the-art methods and demonstrates notable robustness to diverse corruptions and domain shifts.



\title{Supplementary Document: Feature-Weight Cosine Alignment for Test-time Adaptation} 


\author{WeiQin Chuah\orcidlink{0000-0001-6020-4660} \and
Ruwan Tennakoon\orcidlink{0000-0001-8909-5728} \and
Alireza Bab-Hadiashar\orcidlink{0000-0002-6192-2303}}

\authorrunning{W.~Chuah et al.}

\institute{
RMIT University, Melbourne Australia \\ 
\email{\{wei.qin.chuah, ruwan.tennakoon, alireza.bab-hadiashar\}@rmit.edu.au}}

\maketitle

\section{Comprehensive Experimental Results Comparison on ImageNet-C across all five levels of severity.}

As illustrated in Table~\ref{tab:supp-img-rn50} and~\ref{tab:supp-img-rn18}, our proposed CoMM loss function successfully adapts the model across all five levels of severity on the ImageNet-C dataset as well as across all types of corruptions presented in the dataset. For both ResNet-50 and ResNet-18 models, CoMM significantly reduces the average classification error, demonstrating an average error reduction of $19.7\%$ and $18.7\%$, respectively. These results highlight CoMM's comprehensive adaptability and its effectiveness in mitigating the impact of diverse corruption types at varying severity levels, further establishing the method's robust performance in online test-time adaptation scenarios.

\section{Additional Qualitative Analysis}
\begin{figure}[t]
    \centering
    \begin{minipage}{0.05\textwidth}
        \subcaption{} \label{fig:supp_ambiguous_a}
    \end{minipage}%
    \begin{minipage}{0.75\textwidth} 
        \includegraphics[width=\textwidth]{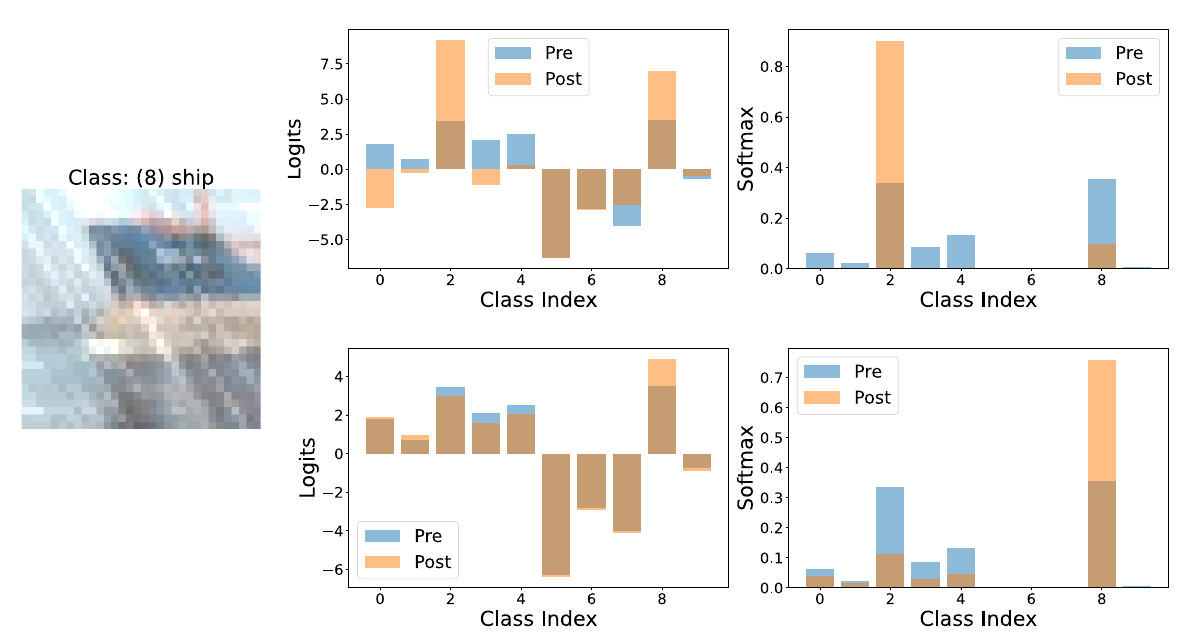}
    \end{minipage}
    \vspace{10pt}
    \begin{minipage}{0.05\textwidth}
        \subcaption{}  \label{fig:supp_ambiguous_b}
    \end{minipage}%
    \begin{minipage}{0.75\textwidth} 
        \includegraphics[width=\textwidth]{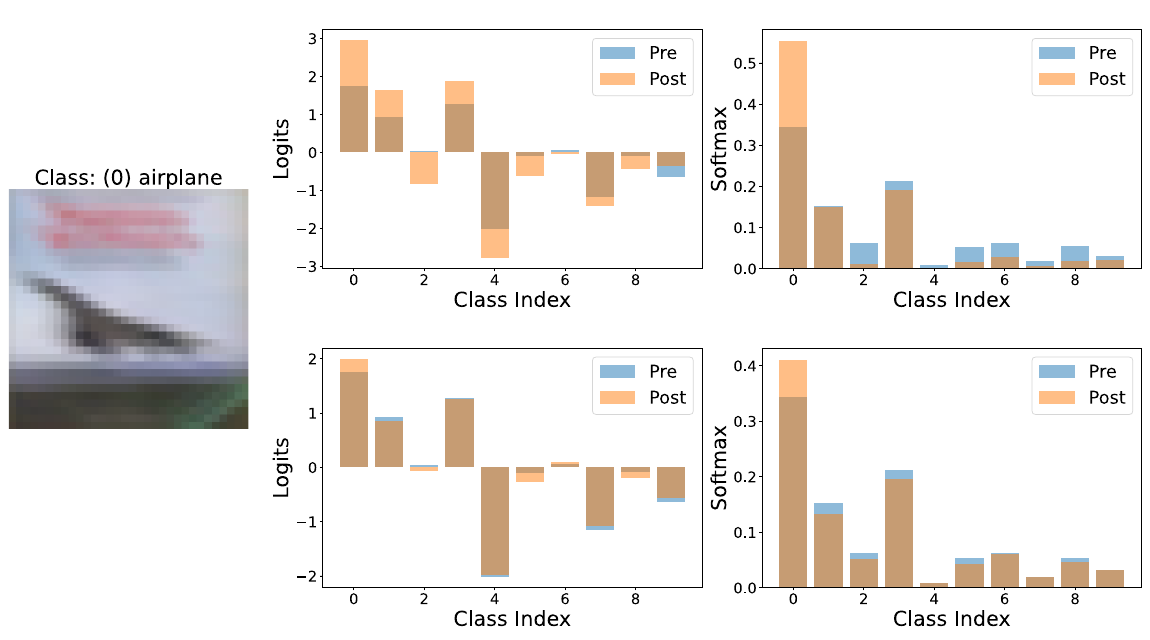}
    \end{minipage}
    \vspace{10pt}
    \begin{minipage}{0.05\textwidth}
        \subcaption{}  \label{fig:supp_ambiguous_c}
    \end{minipage}%
    \begin{minipage}{0.75\textwidth} 
        \includegraphics[width=\textwidth]{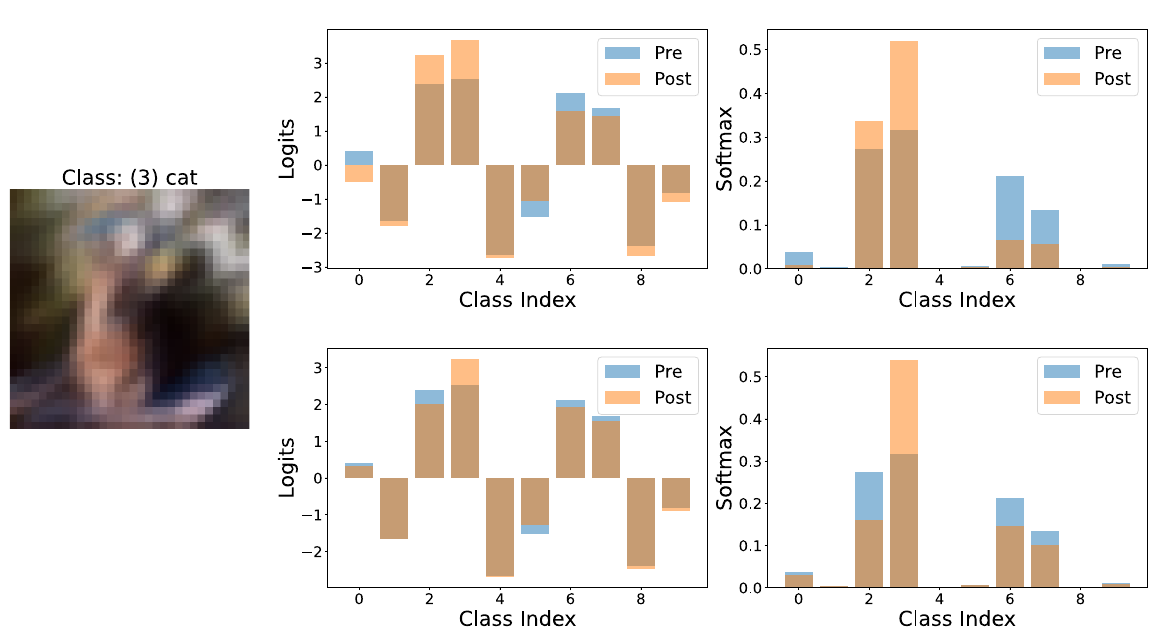}
    \end{minipage}
    \caption{Logit and softmax probability comparisons before and after optimization on CIFAR-10-C samples. Subfigure (a), (b), and (c) showcase the logit adjustments (left) and softmax distributions (right) for three distinct CIFAR-10-C classes—ship, airplane, and cat, respectively. The top row of each panel represents the changes induced by the entropy minimization method, while the bottom row illustrates the adjustments made by our proposed CoMM method, highlighting the differential impact of each method on the model's prediction confidence.}
    \label{fig:supp_ambiguous}
\end{figure}

In Figure~\ref{fig:supp_ambiguous}, we included additional visual examples from CIFAR-10-C to highlight the limitations of the entropy minimization (EM) method~\cite{wang2021tent} in online test-time adaptation on low-margin samples, which was discussed in our main manuscript. Specifically, we compare the changes in logit scores for different classes before and after optimization resulted by EM method and our proposed CoMM. For instance, in Figure~\ref{fig:supp_ambiguous_a}, instead of only enhancing the score for the correct class (ship), the method also increases the score for an incorrect class (cat), indicative of its non-selective optimization. Similar patterns are observed in Figure~\ref{fig:supp_ambiguous_b} and \ref{fig:supp_ambiguous_c}, where the method raises scores for multiple classes, including incorrect ones. This non-selectivity can lead to ambiguous decision-making and less reliable predictions, as the model fails to focus on the most likely class and instead spreads confidence across several classes, potentially diluting the accuracy of the adaptation. 

Conversely, our CoMM method targets the predicted class more precisely, promoting clear and confident decision-making during online adaptation, as evidenced in the corresponding bottom row plots in the figure. As demonstrated by our experimental results included in our main manuscript, this focused adjustment by CoMM is crucial for reliable online adaptation where each prediction impacts subsequent model updates.

\begin{figure}[t]
  \centering
  \begin{subfigure}[b]{0.3\textwidth}
    \includegraphics[width=\textwidth]{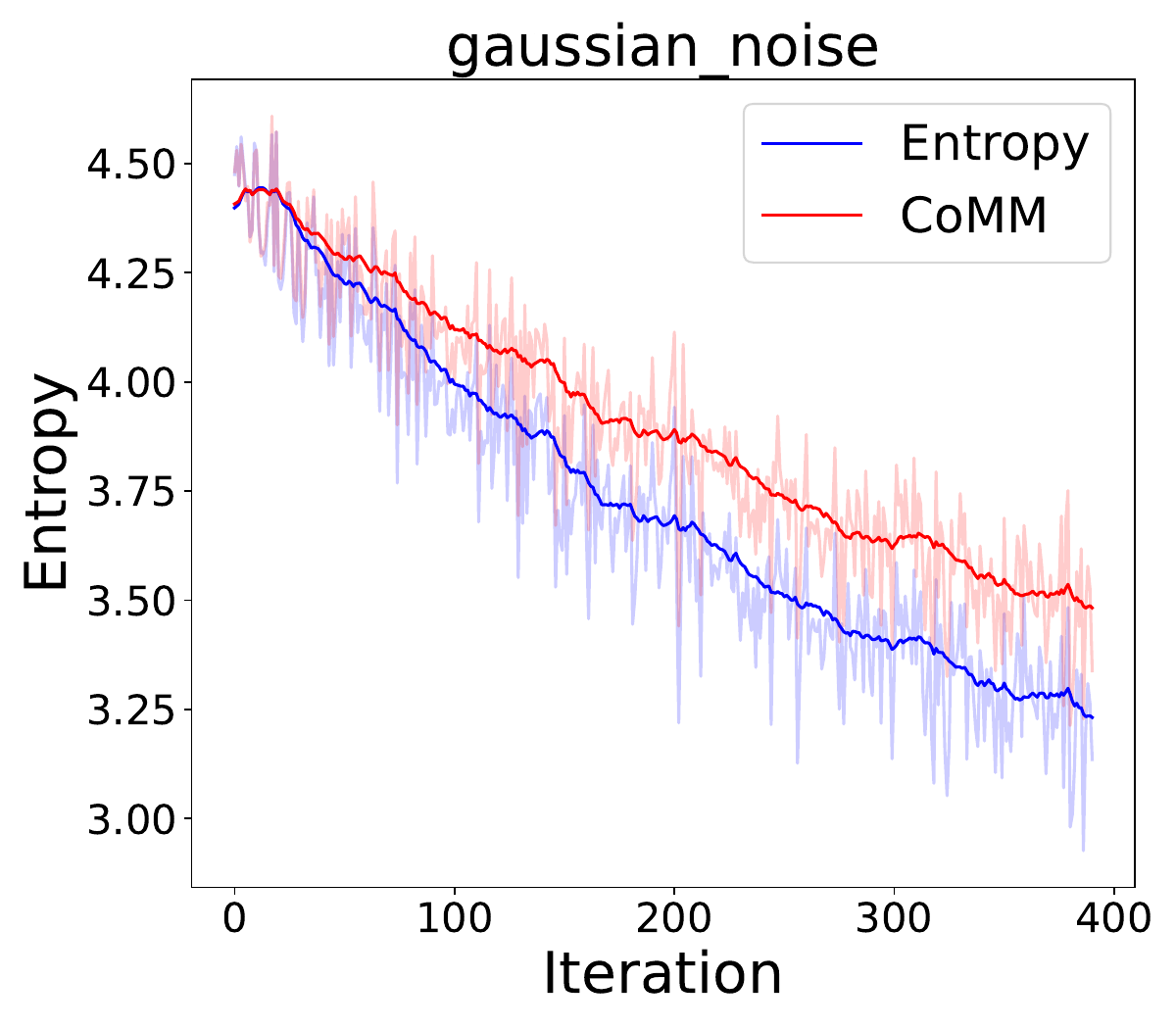}
  \end{subfigure}
  \hfill 
  \begin{subfigure}[b]{0.3\textwidth}
    \includegraphics[width=\textwidth]{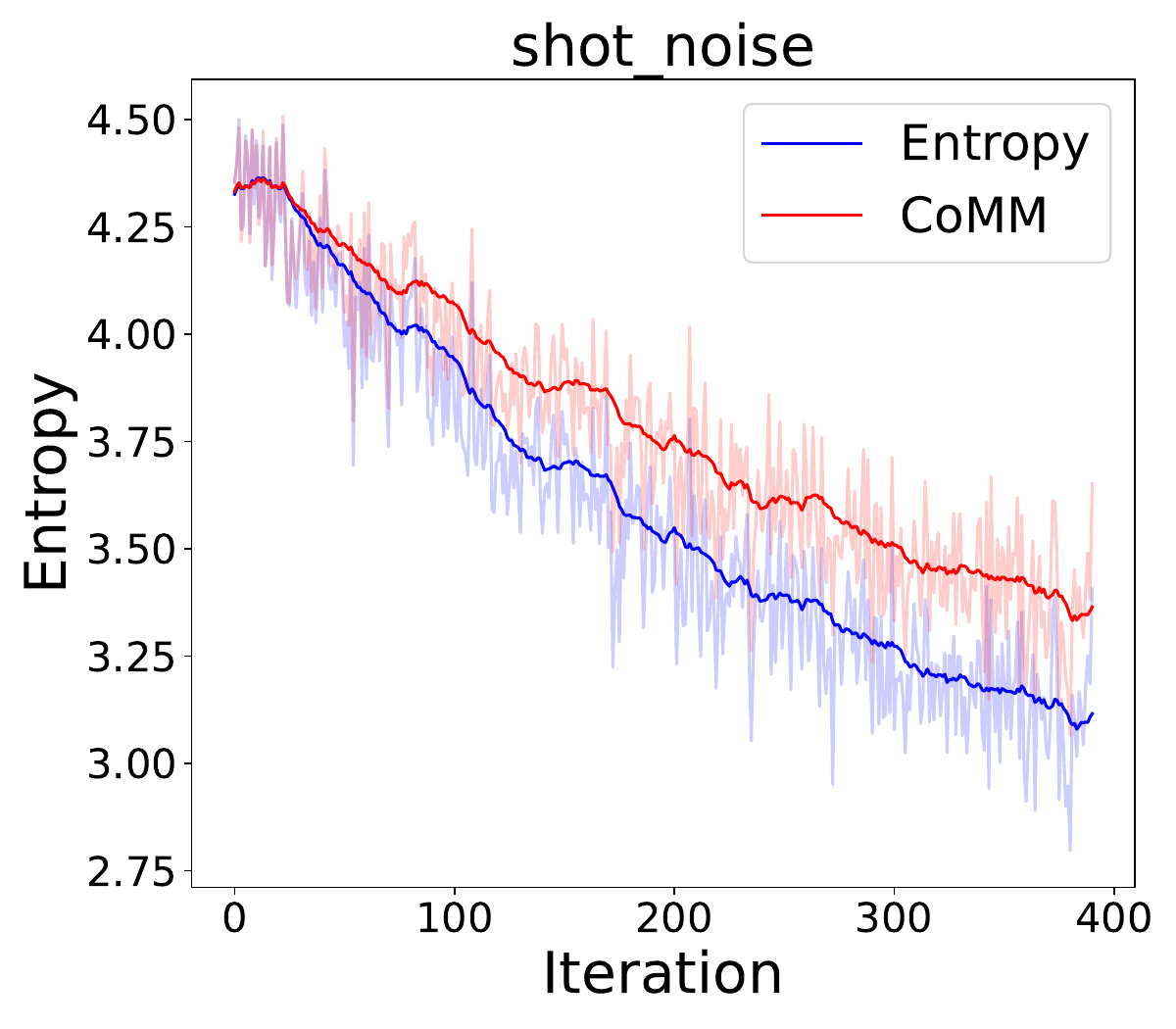}
  \end{subfigure}
  \hfill 
  \begin{subfigure}[b]{0.3\textwidth}
    \includegraphics[width=\textwidth]{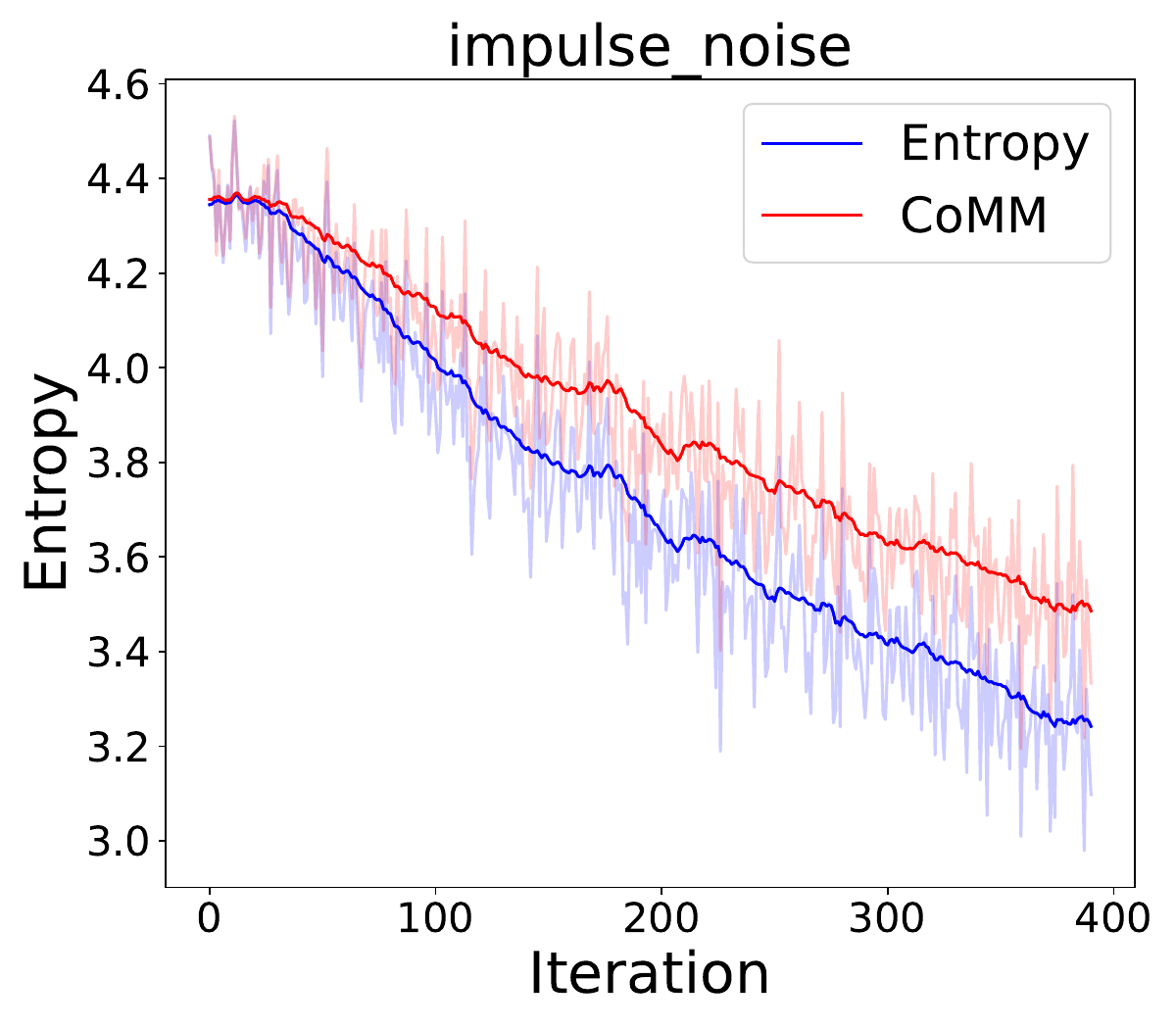}
  \end{subfigure}
  
  \begin{subfigure}[b]{0.3\textwidth}
    \includegraphics[width=\textwidth]{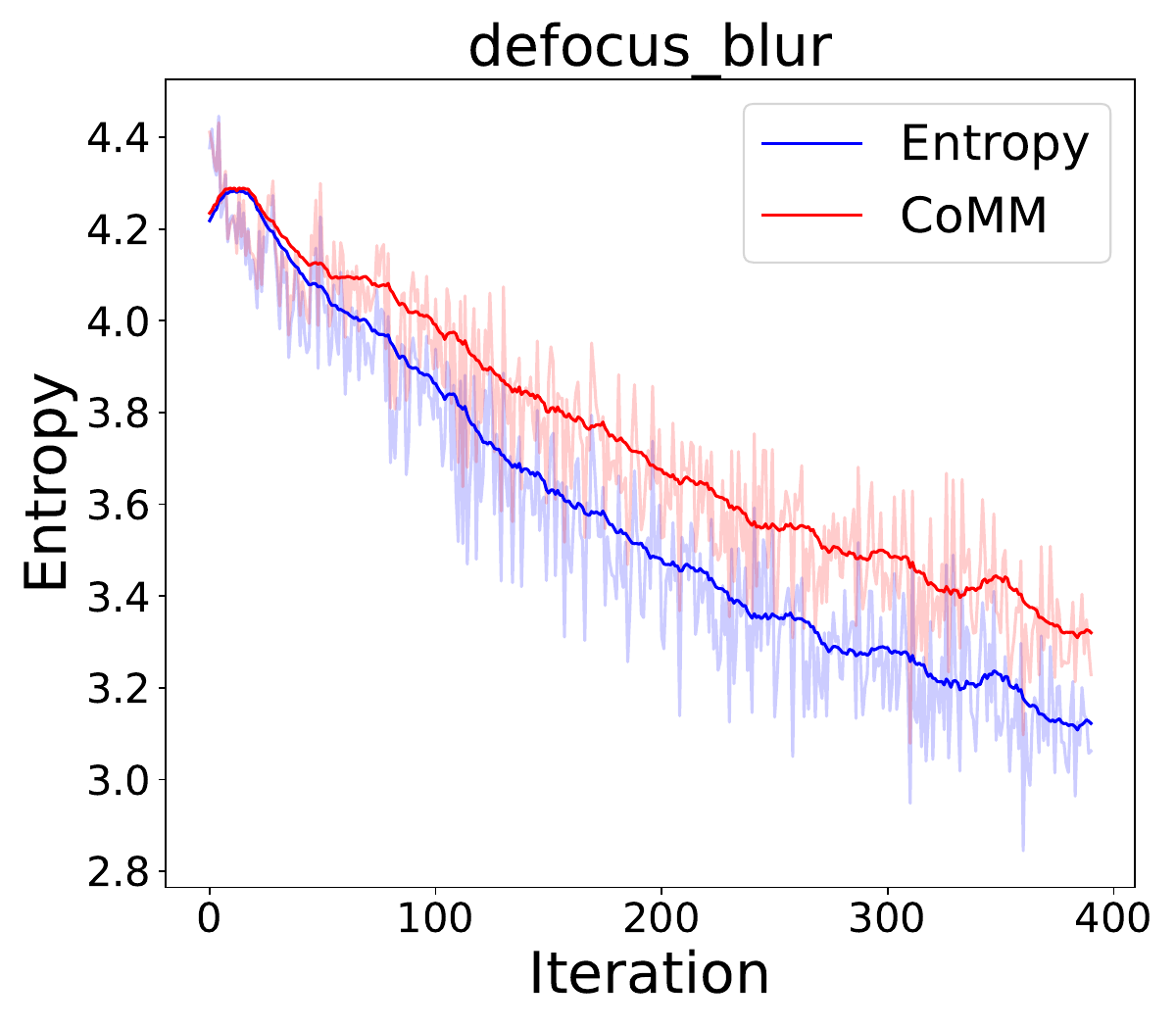}
  \end{subfigure}
  \hfill
  \begin{subfigure}[b]{0.3\textwidth}
    \includegraphics[width=\textwidth]{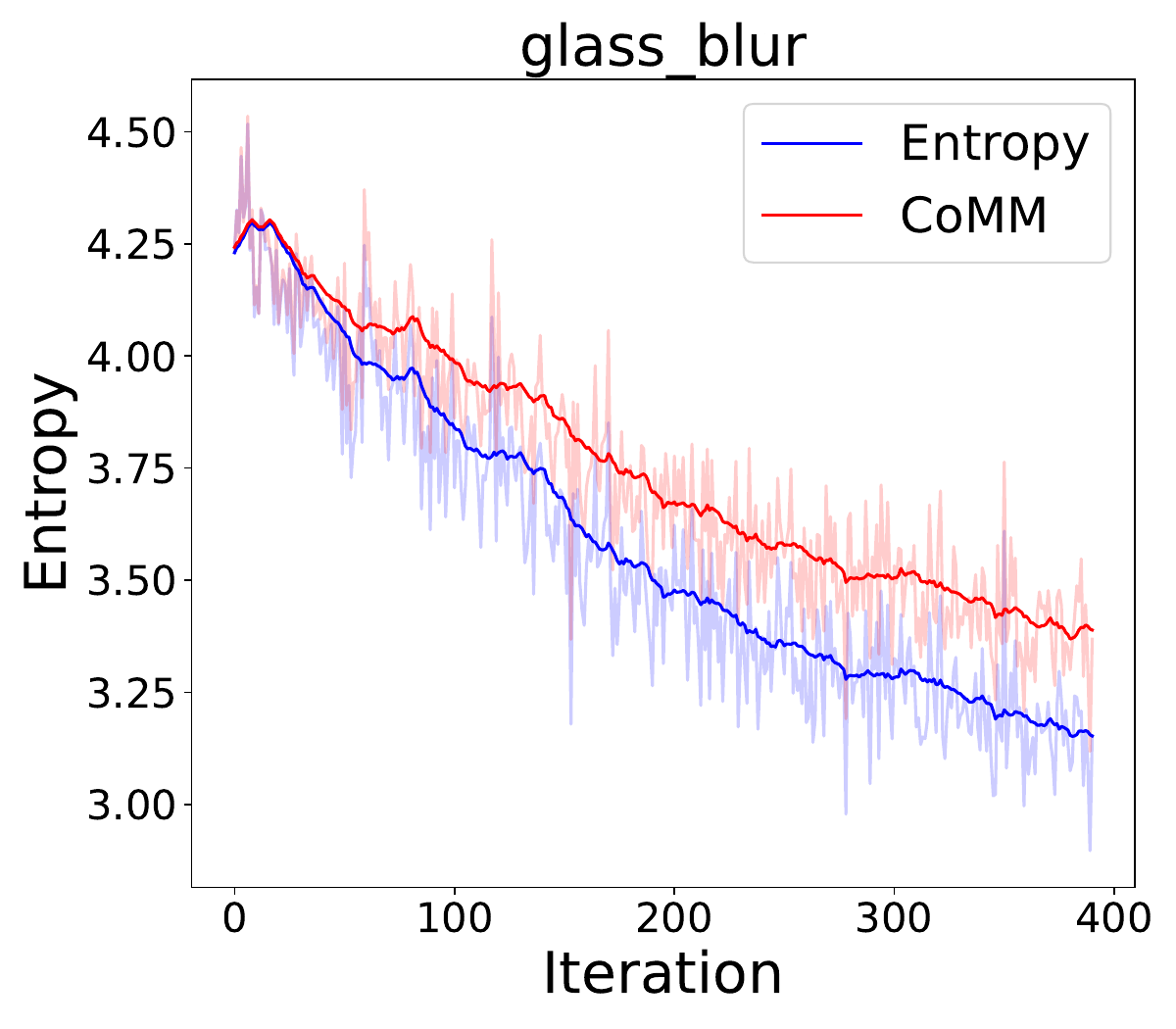}
  \end{subfigure}
  \hfill 
  \begin{subfigure}[b]{0.3\textwidth}
    \includegraphics[width=\textwidth]{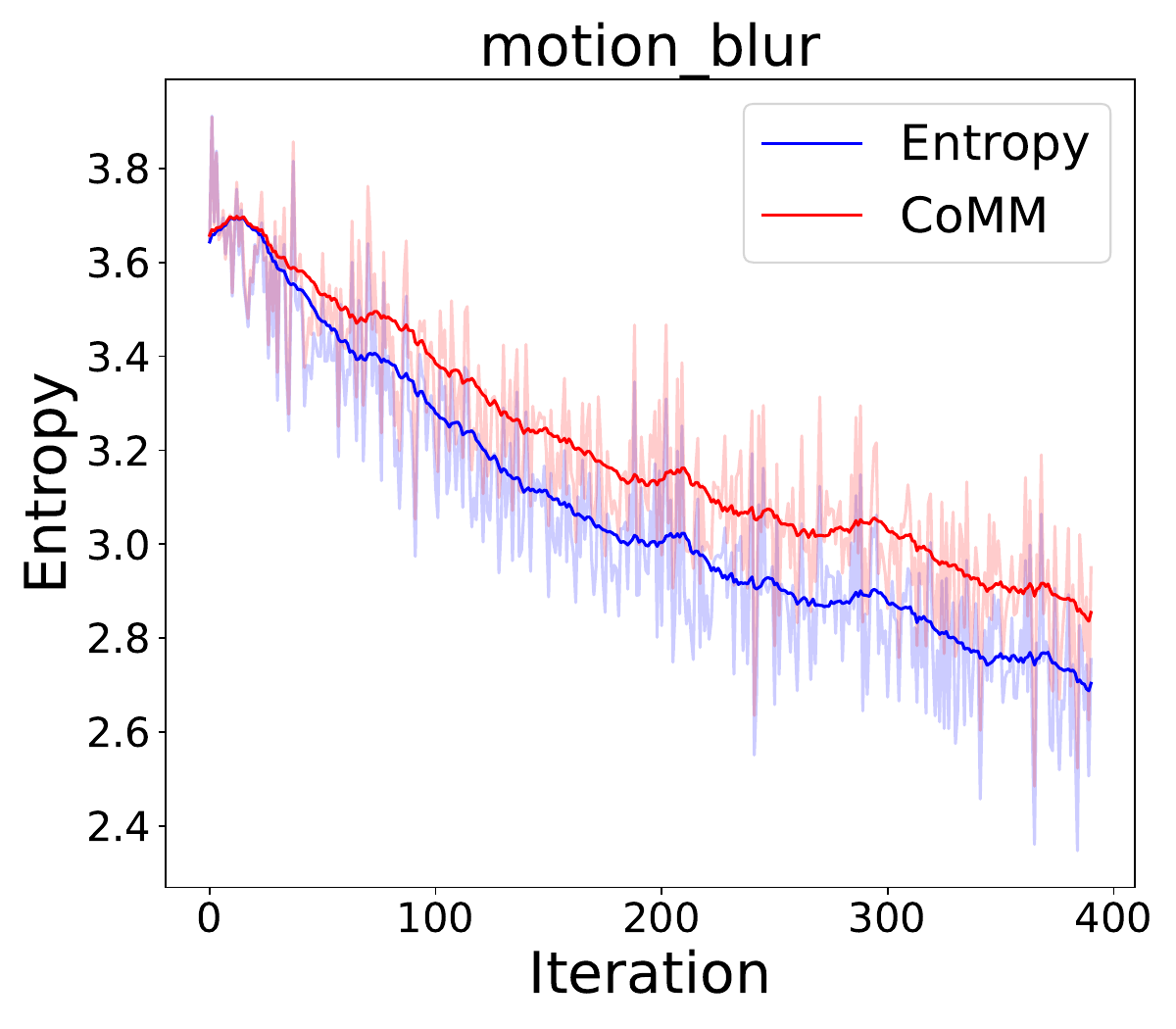}
  \end{subfigure}
  
  \begin{subfigure}[b]{0.3\textwidth}
    \includegraphics[width=\textwidth]{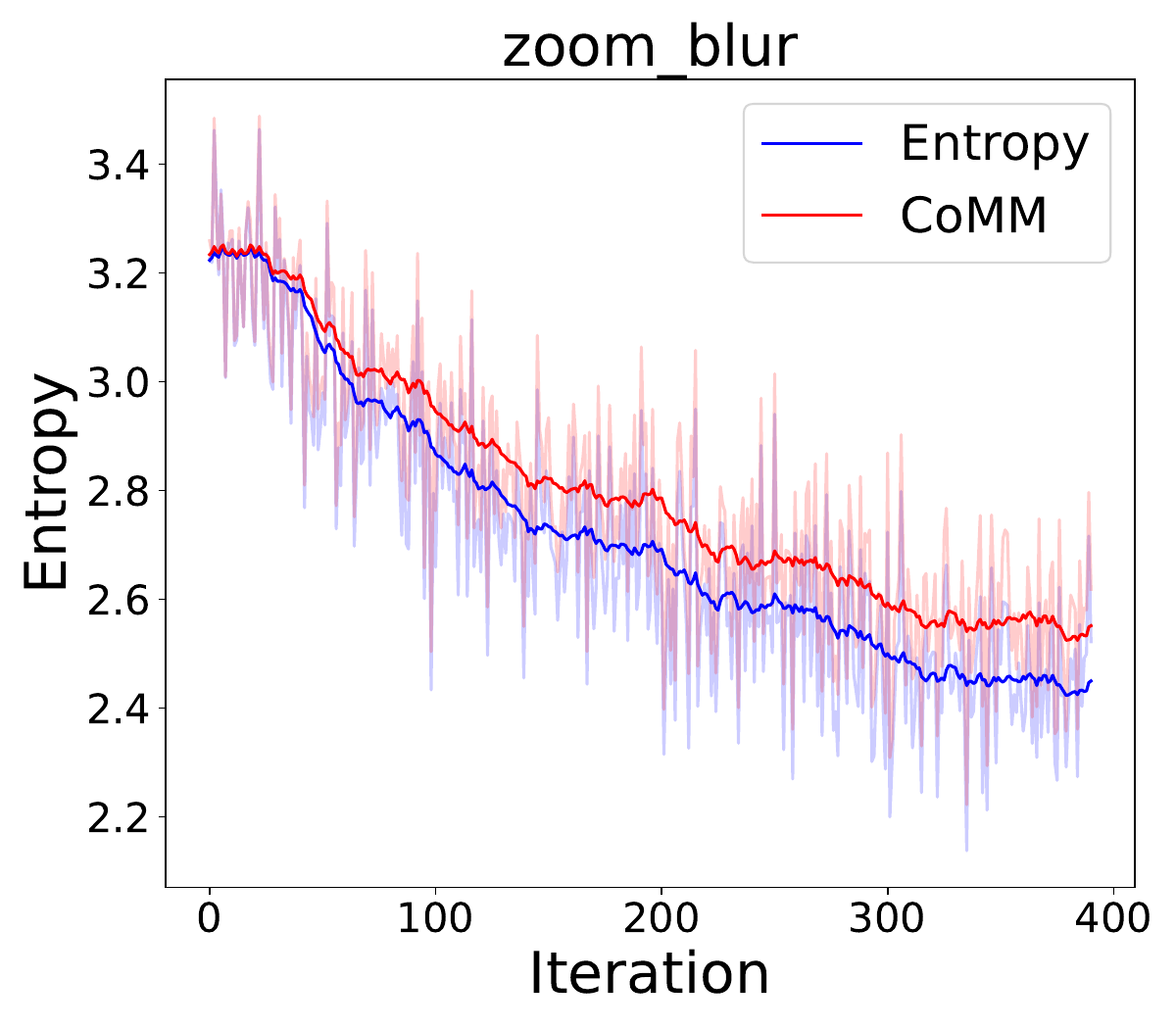}
  \end{subfigure}
  \hfill
  \begin{subfigure}[b]{0.3\textwidth}
    \includegraphics[width=\textwidth]{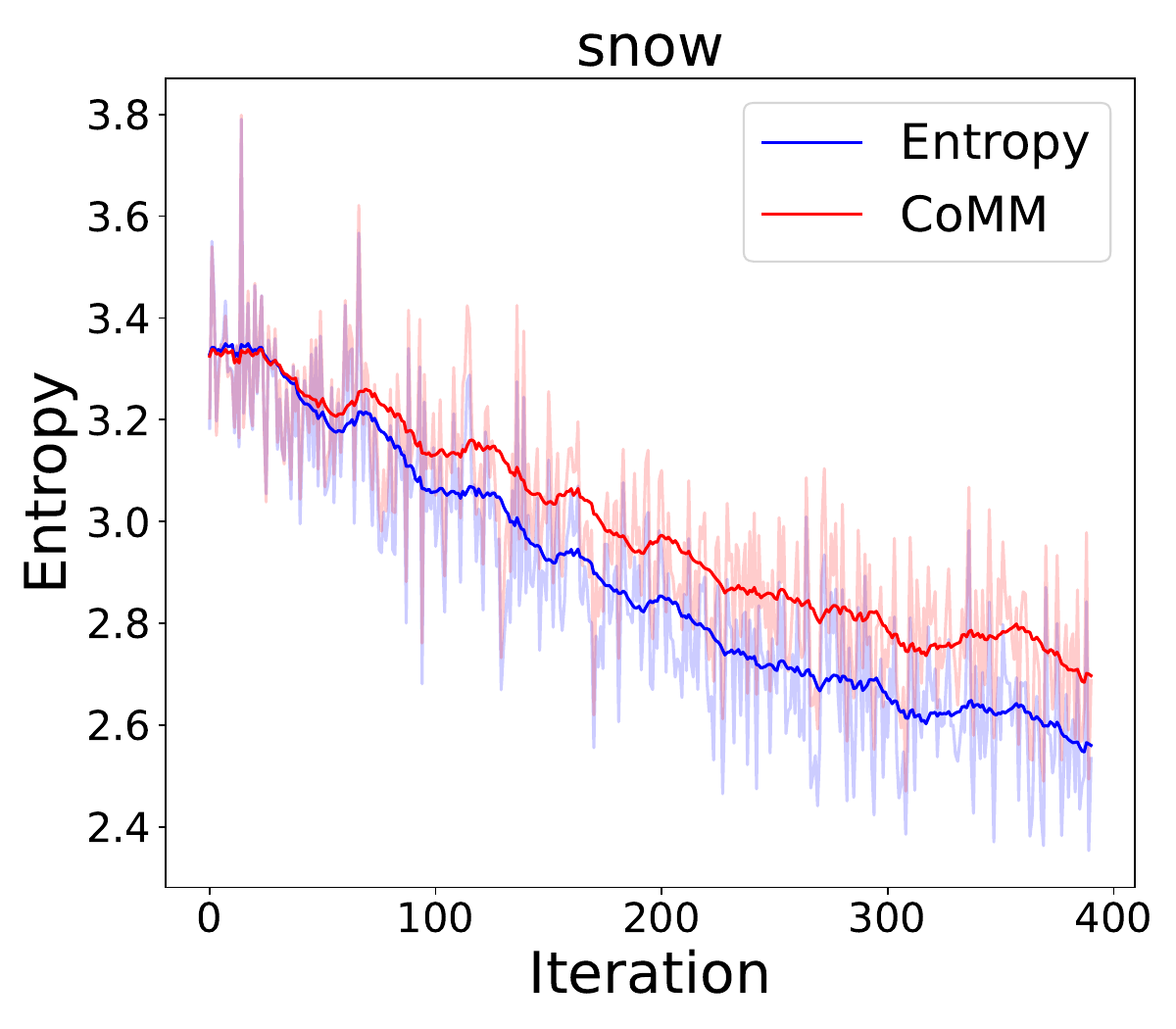}
  \end{subfigure}
  \hfill 
  \begin{subfigure}[b]{0.3\textwidth}
    \includegraphics[width=\textwidth]{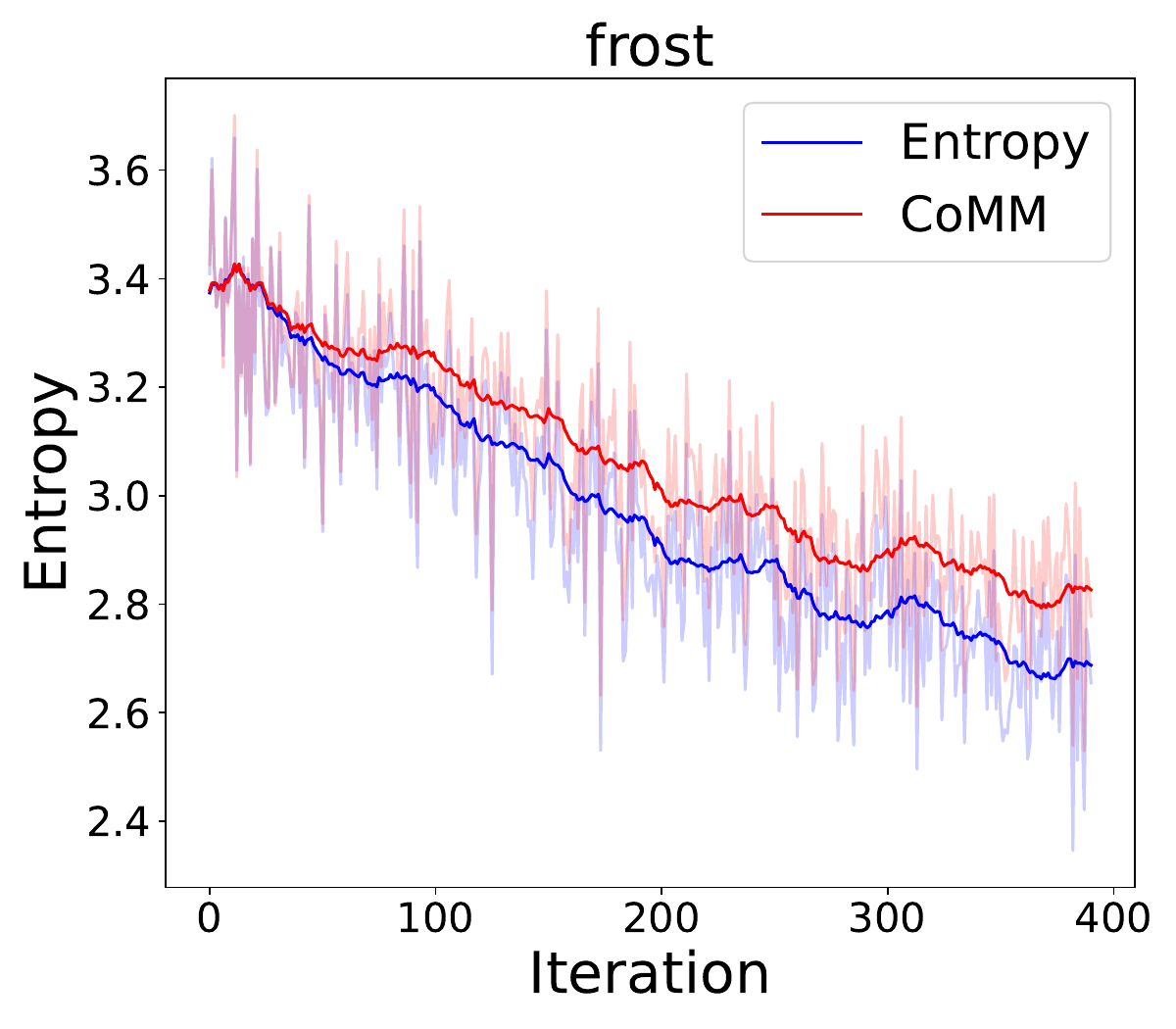}
  \end{subfigure}

  \begin{subfigure}[b]{0.3\textwidth}
    \includegraphics[width=\textwidth]{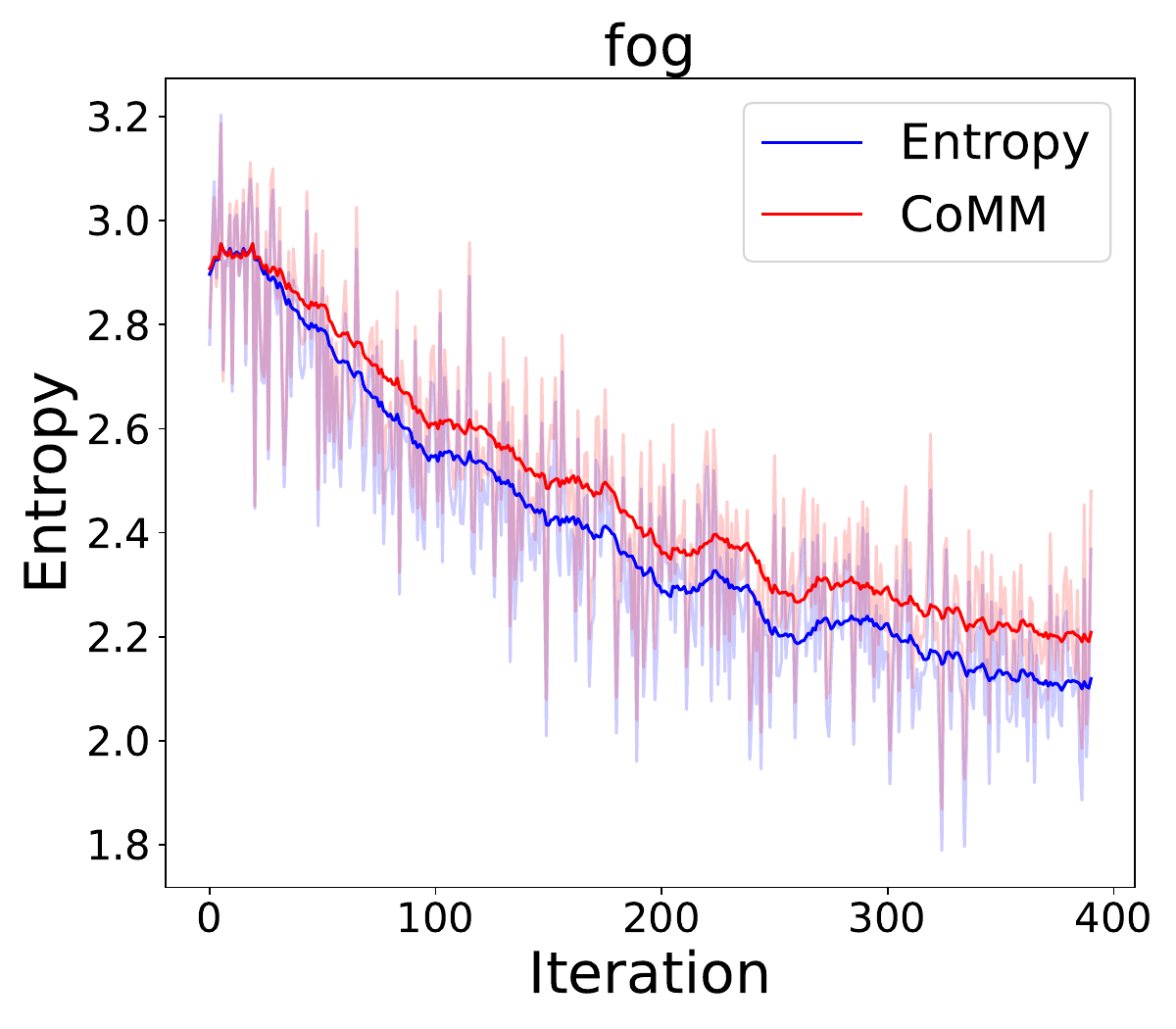}
  \end{subfigure}
  \hfill
  \begin{subfigure}[b]{0.3\textwidth}
    \includegraphics[width=\textwidth]{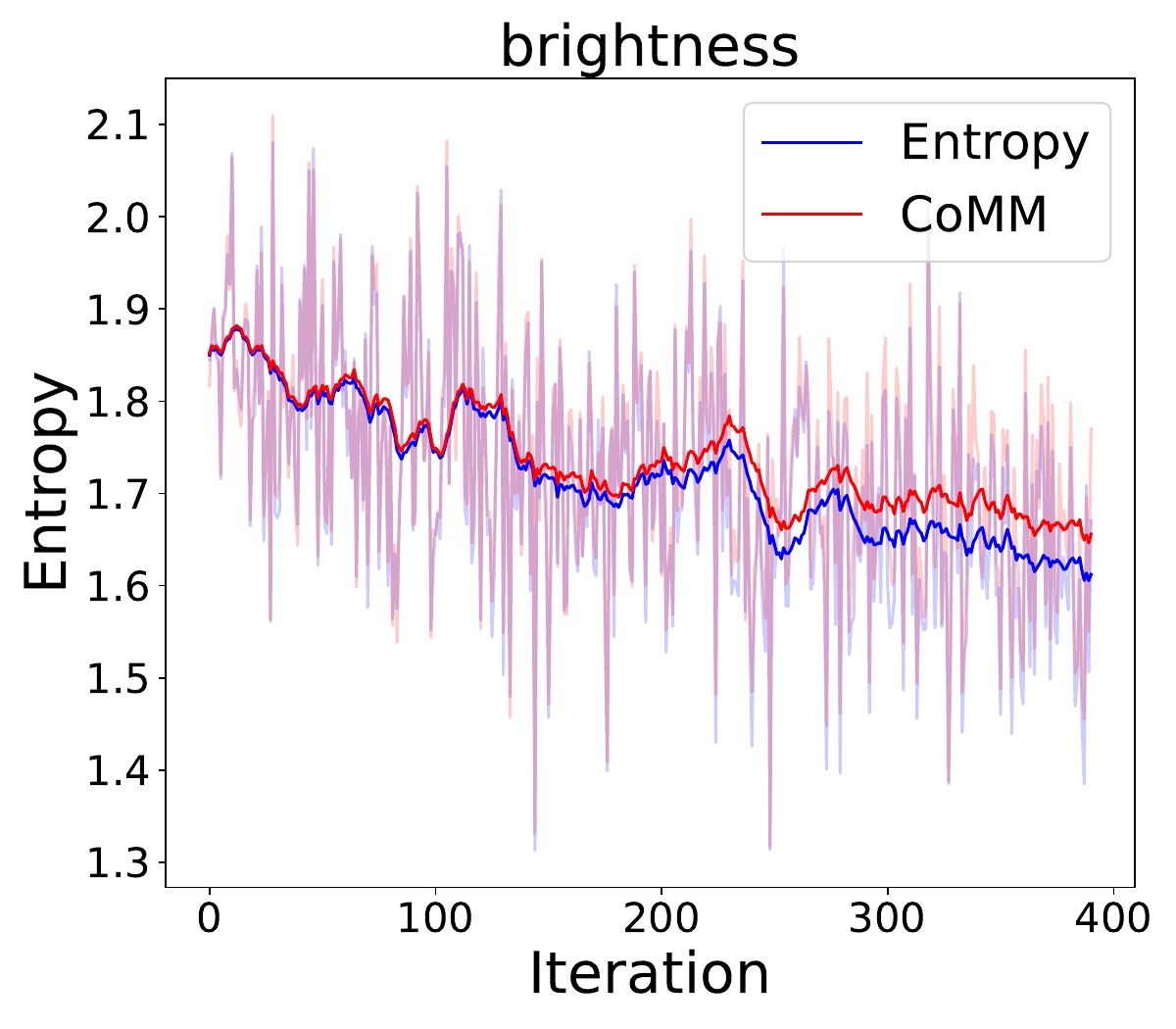}
  \end{subfigure}
  \hfill 
  \begin{subfigure}[b]{0.3\textwidth}
    \includegraphics[width=\textwidth]{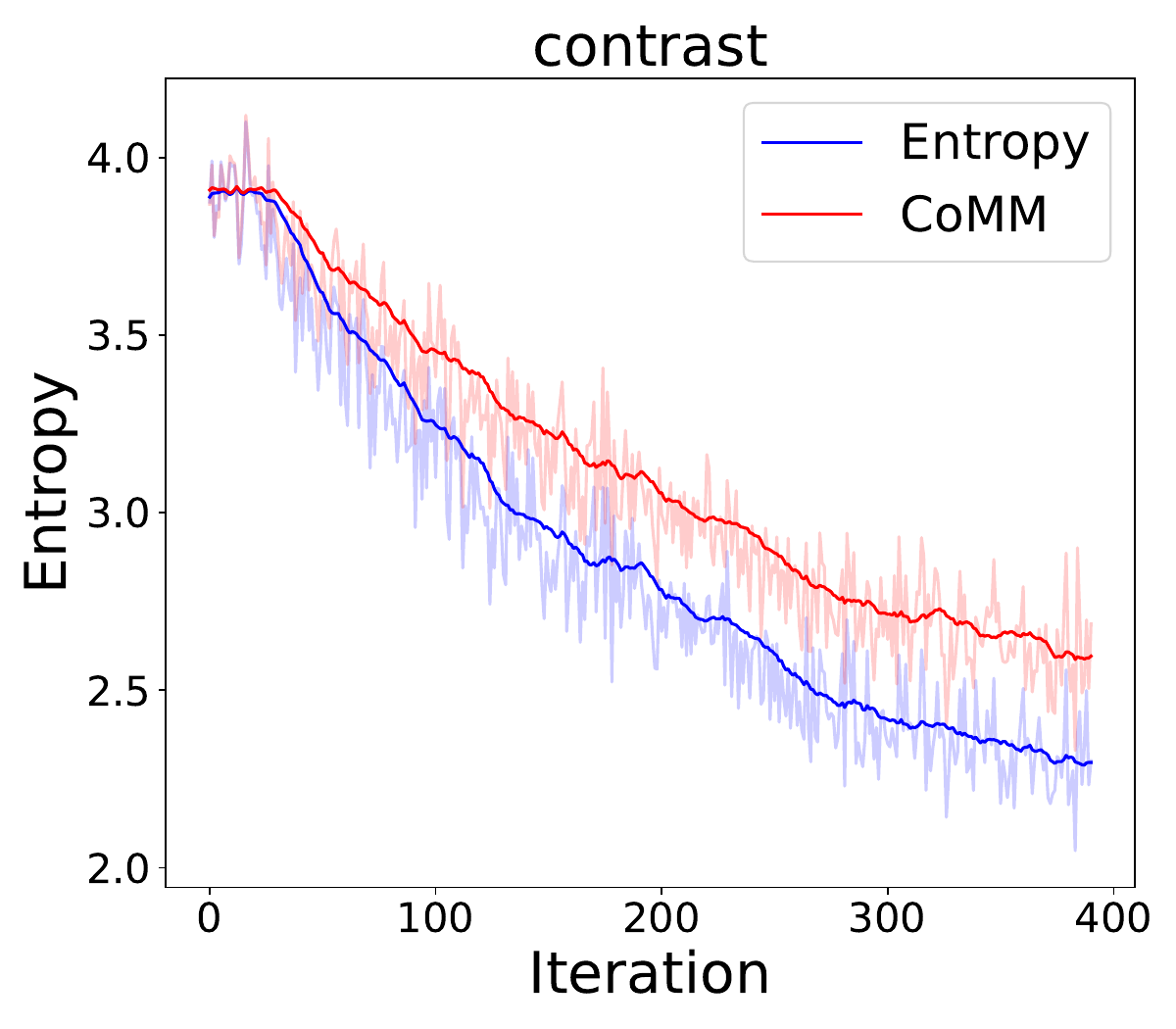}
  \end{subfigure}

  \begin{subfigure}[b]{0.3\textwidth}
    \includegraphics[width=\textwidth]{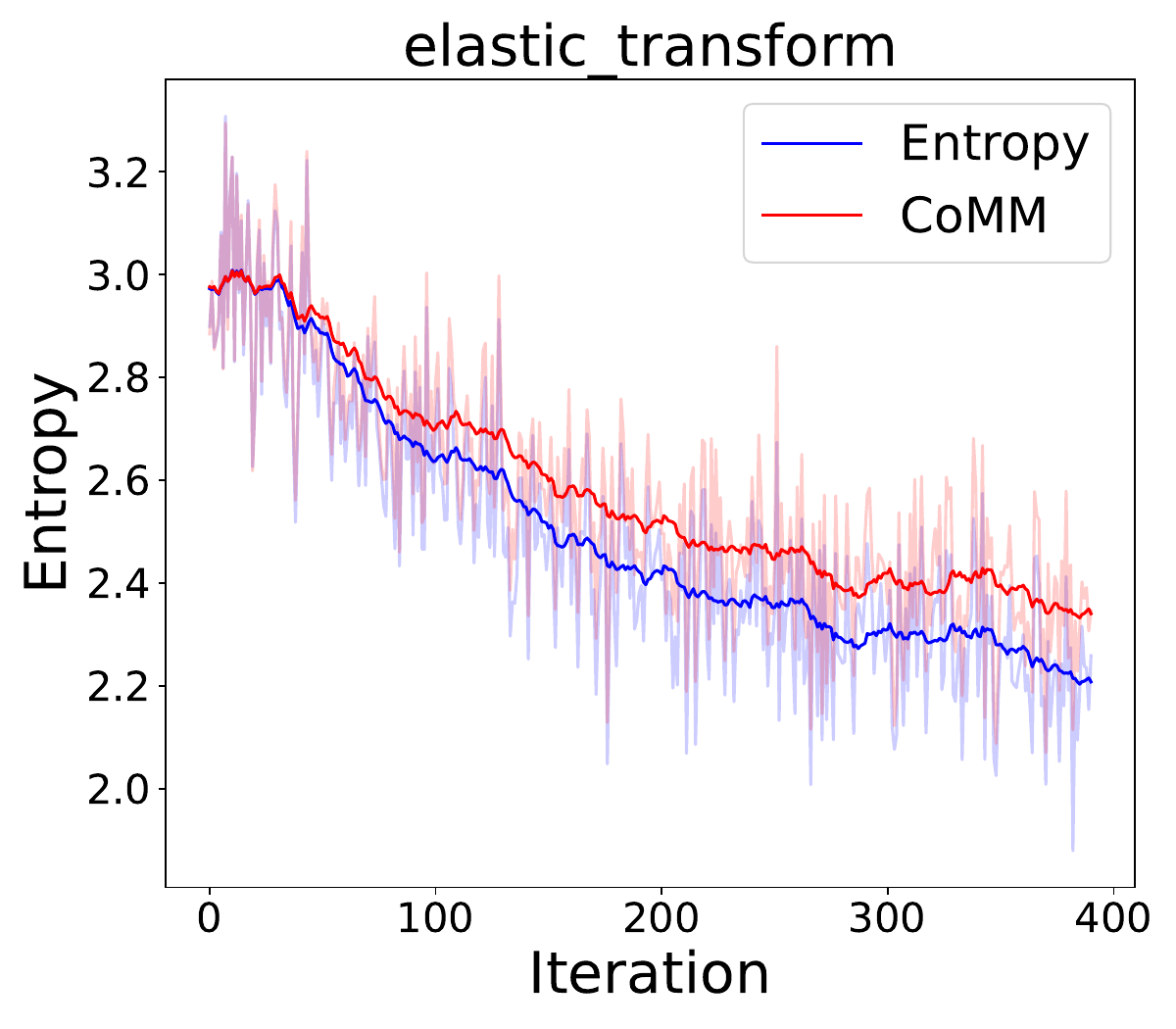}
  \end{subfigure}
  \hfill
  \begin{subfigure}[b]{0.3\textwidth}
    \includegraphics[width=\textwidth]{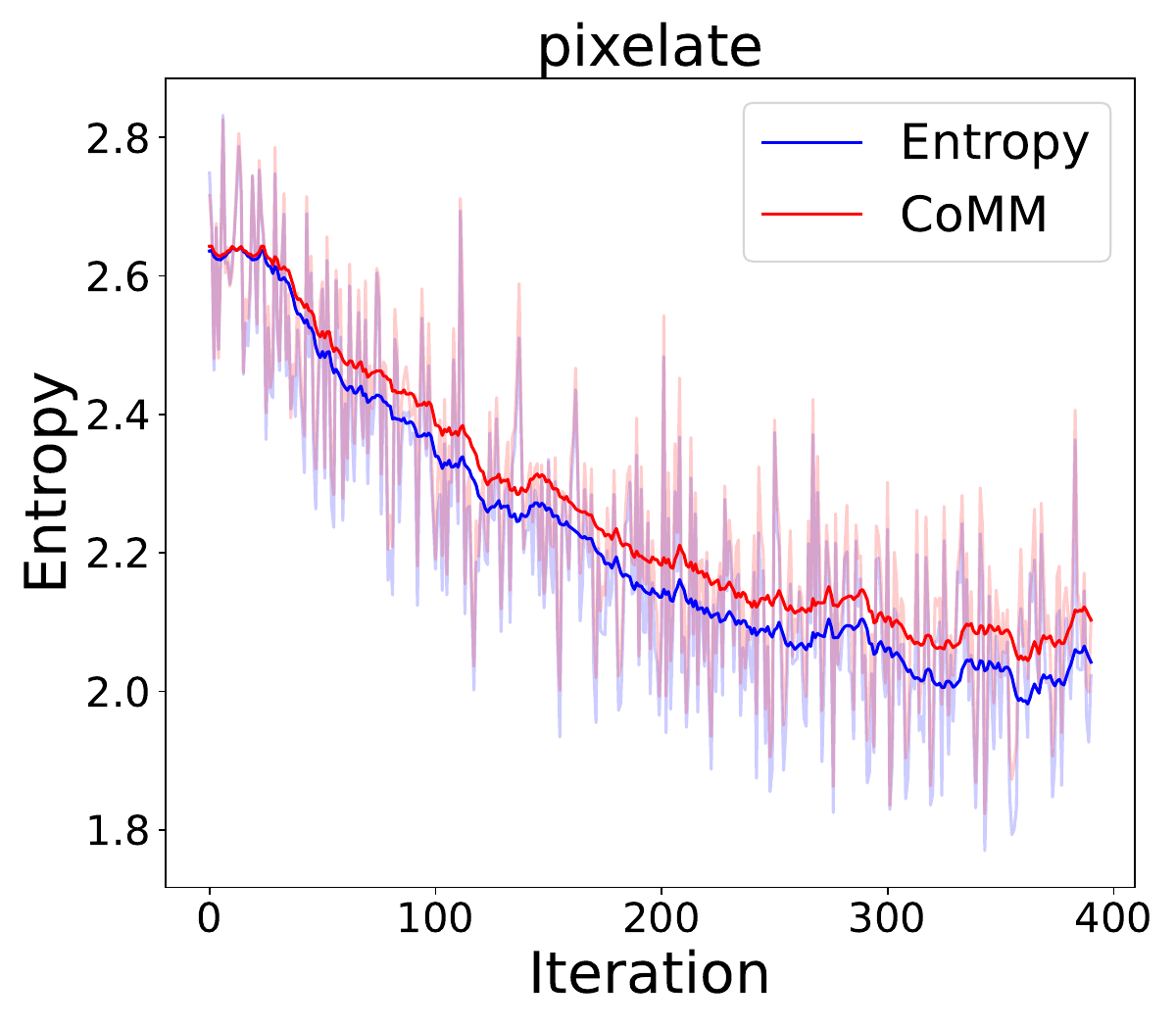}
  \end{subfigure}
  \hfill 
  \begin{subfigure}[b]{0.3\textwidth}
    \includegraphics[width=\textwidth]{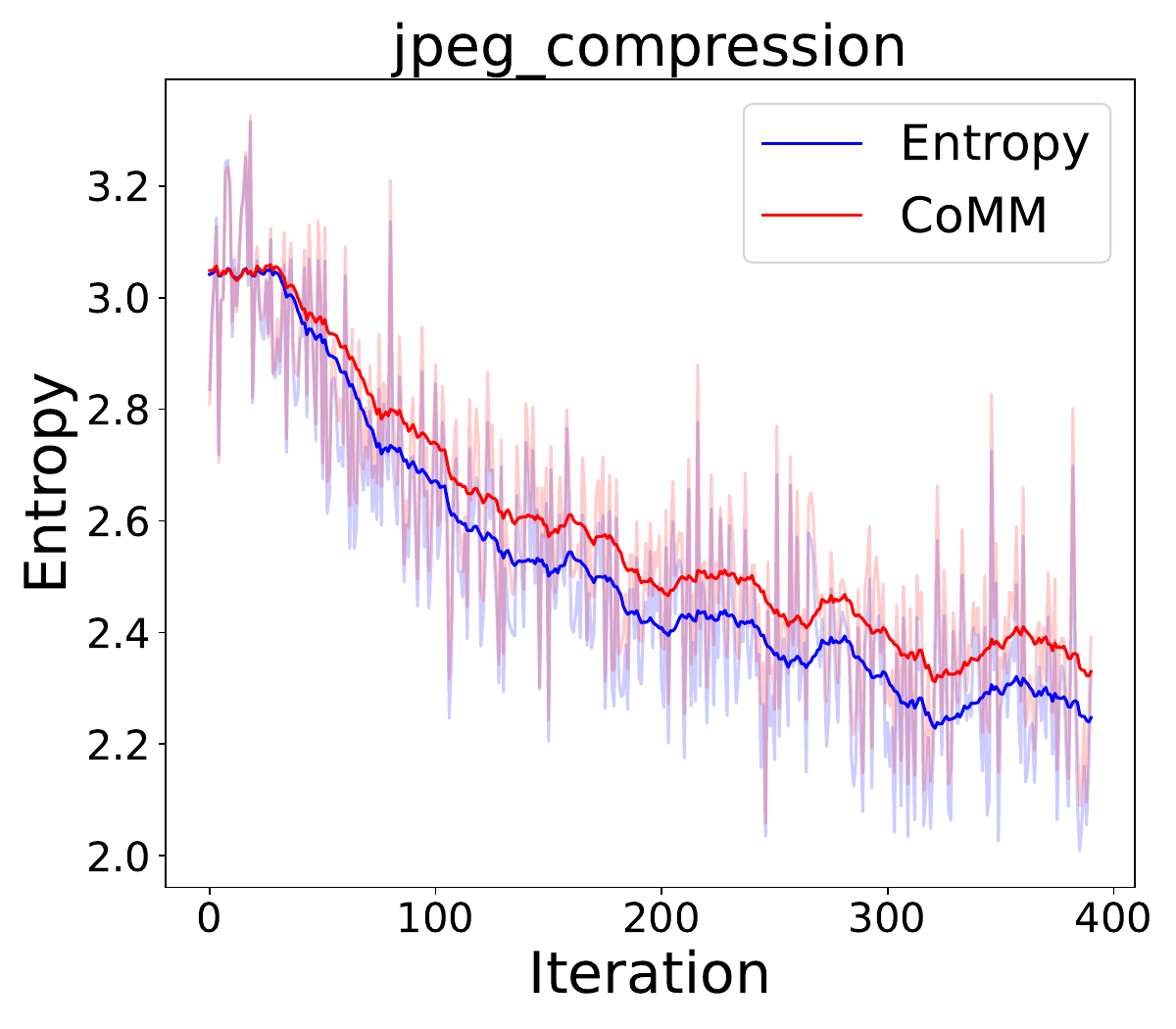}
  \end{subfigure}
  
  \caption{Entropy trends over iteration for Entropy Minimization (blue) and our proposed CoMM (red) across all corruptions on ImageNet-C dataset. While CoMM does not explicitly aim for entropy reduction, these plots show that it naturally decreases prediction entropy over time, suggesting an implicit alignment with the principles of Entropy Minimization. This behavior underlines the ability of CoMM to improve the certainty of the predictions while adapting to out-of-distribution data.}
  \label{fig:entropies}
\end{figure}

In Figure~\ref{fig:entropies}, we compare the model prediction entropy changes during online test-time adaptation. Specifically, we compare the use of entropy minimization (in blue) with our CoMM method (in red) for various ImageNet-C dataset corruptions. Despite that CoMM was not explicitly designed for entropy reduction, it similarly decreases prediction entropy during inference. Notably, CoMM maintains higher entropy levels as compared to entropy minimization, suggesting a potential decrease in overfitting risk by preventing excessive entropy reduction.

\clearpage

\begin{table}[t]
\centering
\caption{comparative analysis of classification error rates on the ImageNet-C dataset using a ResNet-50 model, contrasting the source model with our CoMM loss function across five severity levels.}
\label{tab:supp-img-rn50}
\resizebox{\columnwidth}{!}{%
\begin{tabular}{c@{\hspace{2pt}}|c@{\hspace{2pt}}c@{\hspace{2pt}}c@{\hspace{2pt}}c@{\hspace{2pt}}c@{\hspace{2pt}}c@{\hspace{2pt}}c@{\hspace{2pt}}c@{\hspace{2pt}}c@{\hspace{2pt}}c@{\hspace{2pt}}c@{\hspace{2pt}}c@{\hspace{2pt}}c@{\hspace{2pt}}c@{\hspace{2pt}}c@{\hspace{2pt}}|c}
\hline
\multicolumn{1}{c|}{} &
  \multicolumn{16}{c}{Classification   Error (\%)} \\ \cline{2-17} 
\multicolumn{1}{c|}{\multirow{-2}{*}{Severity}} &
  \textbf{Gaus} &
  \textbf{Shot} &
  \textbf{Impl} &
  \textbf{Dfcs} &
  \textbf{Gls} &
  \textbf{Mtn} &
  \textbf{Zm} &
  \textbf{Snw} &
  \textbf{Frst} &
  \textbf{Fg} &
  \textbf{Brt} &
  \textbf{Cnt} &
  \textbf{Els} &
  \textbf{Px} &
  \multicolumn{1}{c|}{\textbf{Jpg}} &
  \textbf{Mean} \\ \hline
\multicolumn{17}{c}{Source} \\ \hline
\multicolumn{1}{c|}{1} &
  40.6 &
  42.8 &
  52.0 &
  40.6 &
  46.0 &
  35.3 &
  47.6 &
  45.4 &
  38.7 &
  38.2 &
  26.0 &
  35.1 &
  33.3 &
  35.8 &
  \multicolumn{1}{c|}{33.8} &
  39.4 \\
\multicolumn{1}{c|}{2} &
  53.8 &
  57.9 &
  64.2 &
  48.0 &
  59.5 &
  45.7 &
  57.5 &
  68.0 &
  55.9 &
  44.1 &
  27.6 &
  41.6 &
  55.1 &
  35.9 &
  \multicolumn{1}{c|}{7.5} &
  50.2 \\
\multicolumn{1}{c|}{3} &
  72.4 &
  75.0 &
  74.9 &
  62.0 &
  83.1 &
  62.3 &
  64.8 &
  64.7 &
  67.9 &
  53.4 &
  30.4 &
  54.0 &
  44.3 &
  53.8 &
  \multicolumn{1}{c|}{40.7} &
  60.2 \\
\multicolumn{1}{c|}{4} &
  89.0 &
  92.1 &
  91.8 &
  73.5 &
  87.2 &
  78.1 &
  71.6 &
  75.9 &
  70.1 &
  59.6 &
  34.9 &
  79.5 &
  57.8 &
  71.1 &
  \multicolumn{1}{c|}{52.5} &
  72.3 \\
\multicolumn{1}{c|}{5} &
  97.8 &
  97.1 &
  98.2 &
  82.0 &
  90.2 &
  85.3 &
  77.5 &
  83.1 &
  76.7 &
  75.6 &
  41.1 &
  94.6 &
  83.0 &
  79.4 &
  \multicolumn{1}{c|}{68.3} &
  82.0 \\ \hline
\multicolumn{17}{c}{Ours (CoMM)} \\ \hline
\multicolumn{1}{c|}{1} &
  31.7 &
  31.8 &
  35.1 &
  33.1 &
  32.3 &
  29.5 &
  34.2 &
  33.9 &
  33.4 &
  29.7 &
  26.0 &
  28.1 &
  30.2 &
  28.0 &
  \multicolumn{1}{c|}{29.9} &
  31.1 \\
\multicolumn{1}{c|}{2} &
  36.1 &
  36.8 &
  40.1 &
  38.0 &
  37.9 &
  33.2 &
  38.0 &
  42.4 &
  42.2 &
  31.1 &
  26.9 &
  29.5 &
  42.8 &
  28.6 &
  \multicolumn{1}{c|}{31.9} &
  35.7 \\
\multicolumn{1}{c|}{3} &
  43.7 &
  43.3 &
  44.3 &
  48.5 &
  50.3 &
  39.1 &
  40.8 &
  42.2 &
  49.6 &
  33.3 &
  28.1 &
  32.3 &
  30.4 &
  31.9 &
  \multicolumn{1}{c|}{33.4} &
  39.4 \\
\multicolumn{1}{c|}{4} &
  53.8 &
  57.7 &
  54.0 &
  58.4 &
  56.9 &
  47.4 &
  44.5 &
  48.6 &
  50.7 &
  35.2 &
  29.9 &
  40.2 &
  33.1 &
  36.5 &
  \multicolumn{1}{c|}{38.5} &
  45.7 \\
\multicolumn{1}{c|}{5} &
  68.3 &
  68.0 &
  66.5 &
  68.2 &
  69.0 &
  53.9 &
  48.1 &
  49.4 &
  55.8 &
  40.5 &
  32.0 &
  61.6 &
  42.3 &
  39.5 &
  \multicolumn{1}{c|}{45.1} &
  53.9 \\ \hline
\end{tabular}%
}
\end{table}

\begin{table}[t]
\centering
\caption{comparative analysis of classification error rates on the ImageNet-C dataset using a ResNet-18 model, contrasting the source model with our CoMM loss function across five severity levels.}
\label{tab:supp-img-rn18}
\resizebox{\columnwidth}{!}{%
\begin{tabular}
{c@{\hspace{2pt}}|c@{\hspace{2pt}}c@{\hspace{2pt}}c@{\hspace{2pt}}c@{\hspace{2pt}}c@{\hspace{2pt}}c@{\hspace{2pt}}c@{\hspace{2pt}}c@{\hspace{2pt}}c@{\hspace{2pt}}c@{\hspace{2pt}}c@{\hspace{2pt}}c@{\hspace{2pt}}c@{\hspace{2pt}}c@{\hspace{2pt}}c@{\hspace{2pt}}|c}
\hline
\multicolumn{1}{c|}{} &
  \multicolumn{16}{c}{Classification   Error (\%)} \\ \cline{2-17} 
\multicolumn{1}{c|}{\multirow{-2}{*}{Severity}} &
  \textbf{Gaus} &
  \textbf{Shot} &
  \textbf{Impl} &
  \textbf{Dfcs} &
  \textbf{Gls} &
  \textbf{Mtn} &
  \textbf{Zm} &
  \textbf{Snw} &
  \textbf{Frst} &
  \textbf{Fg} &
  \textbf{Brt} &
  \textbf{Cnt} &
  \textbf{Els} &
  \textbf{Px} &
  \multicolumn{1}{c|}{\textbf{Jpg}} &
  \textbf{Mean} \\ \hline
\multicolumn{17}{c}{Source} \\ \hline
\multicolumn{1}{c|}{1} &
  49.7 &
  52.3 &
  65.0 &
  50.1 &
  52.3 &
  43.7 &
  55.3 &
  53.6 &
  48.5 &
  47.3 &
  34.5 &
  42.2 &
  41.7 &
  39.8 &
  \multicolumn{1}{c|}{41.9} &
  47.9 \\
\multicolumn{1}{c|}{2} &
  62.2 &
  66.7 &
  75.9 &
  57.7 &
  64.5 &
  55.6 &
  65.2 &
  76.7 &
  66.1 &
  54.3 &
  36.6 &
  49.1 &
  62.7 &
  39.8 &
  \multicolumn{1}{c|}{45.5} &
  58.6 \\
\multicolumn{1}{c|}{3} &
  80.3 &
  81.9 &
  84.1 &
  72.1 &
  85.2 &
  72.2 &
  71.8 &
  72.4 &
  77.3 &
  64.6 &
  39.7 &
  62.4 &
  49.9 &
  58.3 &
  \multicolumn{1}{c|}{48.4} &
  68.1 \\
\multicolumn{1}{c|}{4} &
  93.5 &
  94.8 &
  94.9 &
  83.4 &
  89.0 &
  84.8 &
  77.3 &
  82.6 &
  79.2 &
  71.8 &
  44.6 &
  86.4 &
  62.5 &
  73.7 &
  \multicolumn{1}{c|}{58.3} &
  78.5 \\
\multicolumn{1}{c|}{5} &
  98.4 &
  97.7 &
  98.4 &
  90.6 &
  92.5 &
  89.8 &
  81.8 &
  89.5 &
  85.0 &
  86.3 &
  51.1 &
  97.2 &
  85.3 &
  76.9 &
  \multicolumn{1}{c|}{71.7} &
  86.2 \\ \hline
\multicolumn{17}{c}{Ours (CoMM)} \\ \hline
\multicolumn{1}{c|}{1} &
  38.6 &
  38.9 &
  43.6 &
  40.3 &
  38.7 &
  35.5 &
  41.0 &
  41.7 &
  40.2 &
  36.1 &
  31.7 &
  34.2 &
  36.6 &
  33.8 &
  \multicolumn{1}{c|}{35.7} &
  37.8 \\
\multicolumn{1}{c|}{2} &
  43.7 &
  44.5 &
  49.1 &
  46.4 &
  45.4 &
  40.1 &
  45.7 &
  52.6 &
  50.5 &
  37.8 &
  32.7 &
  36.0 &
  50.5 &
  34.6 &
  \multicolumn{1}{c|}{37.8} &
  43.2 \\
\multicolumn{1}{c|}{3} &
  51.6 &
  51.6 &
  54.0 &
  58.5 &
  60.9 &
  48.0 &
  49.2 &
  52.3 &
  58.6 &
  40.6 &
  34.3 &
  39.9 &
  36.4 &
  38.4 &
  \multicolumn{1}{c|}{39.5} &
  47.6 \\
\multicolumn{1}{c|}{4} &
  62.7 &
  65.4 &
  65.3 &
  69.9 &
  68.2 &
  59.0 &
  53.3 &
  60.2 &
  59.8 &
  43.0 &
  36.9 &
  52.1 &
  39.9 &
  44.5 &
  \multicolumn{1}{c|}{45.2} &
  55.0 \\
\multicolumn{1}{c|}{5} &
  76.3 &
  73.6 &
  76.3 &
  76.1 &
  77.3 &
  59.9 &
  53.7 &
  56.5 &
  65.6 &
  47.1 &
  38.4 &
  87.7 &
  49.3 &
  46.0 &
  \multicolumn{1}{c|}{51.1} &
  62.3 \\ \hline
\end{tabular}%
}
\end{table}

\clearpage
%
%
\bibliographystyle{splncs04}
\bibliography{egbib}


%
%
\end{document}